\documentclass[runningheads]{llncs}
\usepackage{hyperref}
\hypersetup{ colorlinks,
linkcolor=red,
urlcolor=magenta,}
\usepackage[T1]{fontenc}
\usepackage{graphicx}
\usepackage{multirow}
\usepackage[dvipsnames]{xcolor}
\usepackage{subcaption}
\usepackage{amssymb}
\usepackage{pifont}
\newcommand{\cmark}{\ding{51}}
\newcommand{\xmark}{\ding{55}}
\begin{document}
\title{GenFormer -- Generated Images are All You Need to Improve Robustness of Transformers on Small Datasets}
\titlerunning{GenFormer}

\author{Sven Oehri\inst{1,*} \and
Nikolas Ebert\inst{1,2,*}
\and
Ahmed Abdullah\inst{1} \and
Didier Stricker\inst{2} \and
Oliver Wasenmüller\inst{1}}
\authorrunning{S. Oehri, N. Ebert et al.}

\institute{\textsuperscript{1}Mannheim University of Applied Sciences, Germany \\
\textsuperscript{2}University of Kaiserslautern-Landau (RPTU), Germany \\
\email{\{s.oehri,n.ebert,a.abdullah,o.wasenmueller\}@hs-mannheim.de} \space
\email{didier.stricker@dfki.de}}

% %
\maketitle              % typeset the header of the contribution
\def\thefootnote{*}\footnotetext{equal contribution}\def\thefootnote{\arabic{footnote}}
\begin{abstract}
Recent studies showcase the competitive accuracy of Vision Transformers (ViTs) in relation to Convolutional Neural Networks (CNNs), along with their remarkable robustness.
However, ViTs demand a large amount of data to achieve adequate performance, which makes their application to small datasets challenging, falling behind CNNs.
To overcome this, we propose GenFormer, a data augmentation strategy utilizing generated images, thereby improving transformer accuracy and robustness on small-scale image classification tasks.
In our comprehensive evaluation we propose Tiny ImageNetV2, -R, and -A as new test set variants of Tiny ImageNet by transferring established ImageNet generalization and robustness benchmarks to the small-scale data domain.
Similarly, we introduce MedMNIST-C and EuroSAT-C as corrupted test set variants of established fine-grained datasets in the medical and aerial domain.
Through a series of experiments conducted on small datasets of various domains, including Tiny ImageNet, CIFAR, EuroSAT and MedMNIST datasets, we demonstrate the synergistic power of our method, in particular when combined with common train and test time augmentations, knowledge distillation, and architectural design choices. 
Additionally, we prove the effectiveness of our approach under challenging conditions with limited training data, demonstrating significant improvements in both accuracy and robustness, bridging the gap between CNNs and ViTs in the small-scale dataset domain.

\end{abstract}    
\section{Introduction}
\label{sec:intro}

\begin{figure}[t]
\centering
\includegraphics[width=0.99\linewidth]{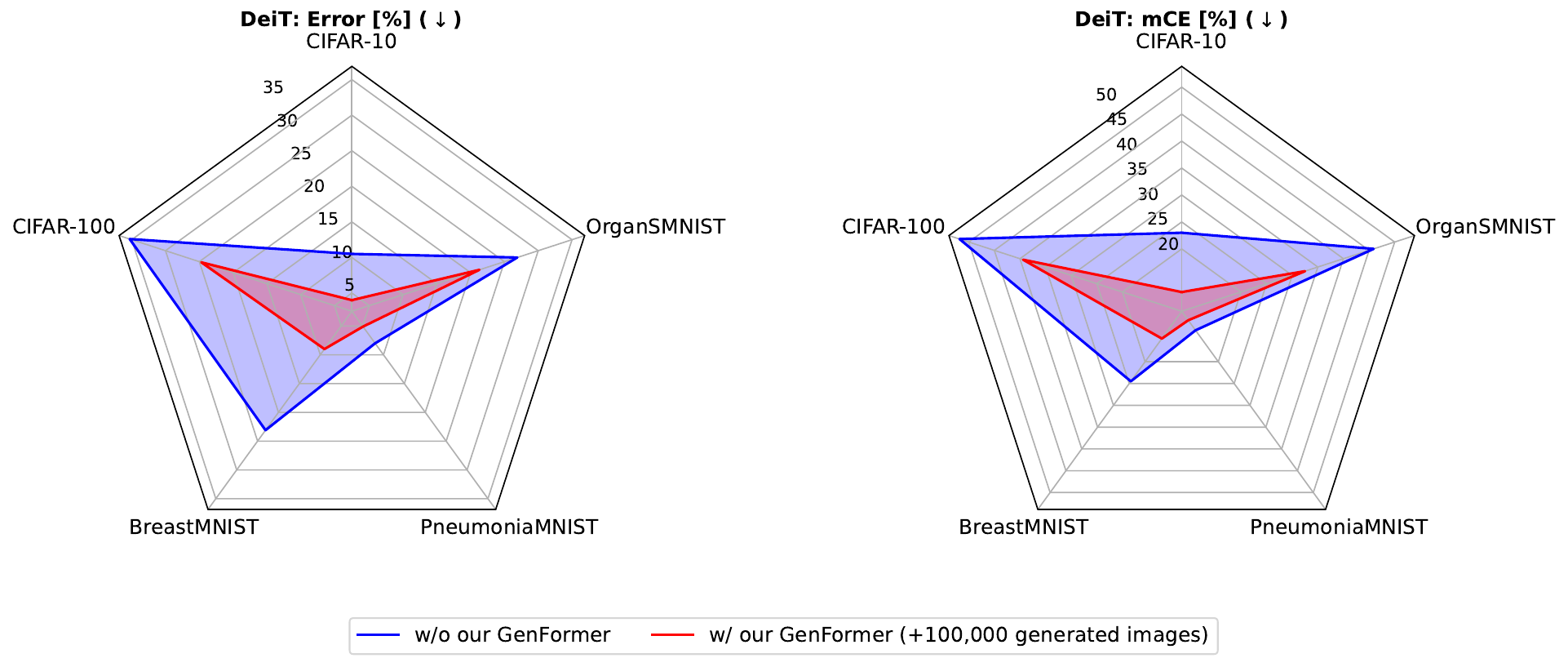} 
\caption{Comparison of the error rate (left) and mean corruption error (right) of DeiT \cite{touvron2021training} on CIFAR \cite{krizhevsky2009learning} and parts of the MedMNIST \cite{medmnistv2} collection \textcolor{red}{with} and \textcolor{blue}{without} our GenFormer. Lower error rates closer to the plot center are better.
}

\label{fig:c100_summary}
\vspace{-6mm}
\end{figure}

Deep learning models, whether based on convolution or self-attention, achieve remarkable performances across a wide range of computer vision benchmarks.
Yet, numerous works demonstrate the vulnerability of modern architectures to adversarial perturbations \cite{lovisotto2022give},
%\cite{fu2021patch,gu2022vision,madry2018towards}
common corruptions \cite{hendrycks2018benchmarking}, 
and domain shifts \cite{hendrycks2021many},
a major challenge on the road to real-world applications. 
Recent research demonstrates the intrinsic robustness and generalization superiority of the transformer architecture compared to Convolutional Neural Networks (CNNs) \cite{zhang2022delving,bhojanapalli2021understanding,paul2022vision}.  % \cite{bhojanapalli2021understanding,paul2022vision}. %removed: zhang2022delving
However, a huge drawback of the Vision Transformer (ViT) \cite{dosovitskiy2020image} and its variants \cite{d2021convit,ebert2023light,ebert2023plg,wang2022pvt} remains the demand for large-scale training data, due to its lack of an inductive bias which makes ViTs prone to overfitting when data is scarce.%, as can be seen in Figure \ref{fig:overfit} (blue line).

Transfer learning is a commonly chosen approach to mitigate the problem of data scarcity by pre-training the model on an out-of-domain medium- or large-scale dataset \cite{deng2009imagenet} and subsequently fine-tuning on the target data.
However, the viability of this approach is limited since specific domains (e.g. medical imaging \cite{medmnistv2}) lack large-scale datasets.
% Moreover, certain scenarios would require a repeat of pre-training due to architectural changes necessary for specific tasks, resulting in expensive computational costs. %removed: zhu2021hard qiao2021detectors li2018detnet
Similar constraints apply to self-supervised approaches, such as masked image modeling (MIM) \cite{he2022masked} which has proven to also rely on large amounts of data \cite{xie2023data}.
% Optional citings for MIM: \cite{doersch2015unsupervised, chen2020generative}
Thus, a plethora of works propose methods enabling ViTs to be trained on medium and small-sized datasets without the use of extra data \cite{lee2021vision,liu2021efficient,touvron2021training}. 
%However, exploiting the inherent robustness of transformers in limited-data scenarios remains a sparsely explored field.
However, exploiting the inherent robustness of transformers in data-limited scenarios remains a sparsely explored field.

In this work, we propose GenFormer, a data augmentation scheme enhancing the applicability of ViTs to small-scale datasets, utilizing their inherent robustness, by directly tackling data scarcity. 
We obtain additional information from the training data by expanding the real dataset with images from a generative model. Our GenFormer approach showcases accuracy and robustness improvements for a variety of transformer networks on downstream classification tasks, especially in domains with limited access to data (e.g. medical imaging \cite{medmnistv2}), as demonstrated by the results of DeiT \cite{touvron2021training} in Figure \ref{fig:c100_summary}. 
Since this work aims for a comprehensive investigation of our method's impact on robustness in limited-data scenarios, we propose Tiny ImageNetV2, -R, and -A as new test sets of Tiny ImageNet by transferring established ImageNet \cite{deng2009imagenet} generalization \cite{recht2019imagenet} and robustness \cite{hendrycks2021many,hendrycks2021nae} benchmarks to the small-scale data domain. 
Furthermore, we demonstrate the straightforward applicability and synergistic effectiveness of our training scheme by combining it with existing approaches incorporating train and test time augmentation, knowledge distillation and architectural adaptions.

Our code and models are available on GitHub at \url{https://github.com/CeMOS-IS/GenFormer}. All test sets for evaluation can be found at \url{https://github.com/CeMOS-IS/Robust-Minisets}.

\section{Related Works}
\label{sec:rw}

\textbf{Vision Transformers for Small Datasets.}
Recent efforts enable the use of Vision Transformers (ViTs) on small-scale datasets through novel training approaches. Strategies include advanced data augmentation \cite{touvron2021training,lee2021vision}, knowledge distillation from CNNs \cite{touvron2021training,li2022locality}, and self-supervised objectives promoting spatial understanding \cite{liu2021efficient}. Further approaches introduce architectural adaptations, such as adding convolutions to induce locality \cite{lu2022bridging}, %hassani2021escaping
or explicitly focusing on adapting the self-attention module to encourage a stronger focus on local \cite{lee2021vision} and meaningful \cite{lu2022bridging,chefer2022optimizing} 
information to prevent performance degradation on scarce datasets. 

\textbf{Robustness of Vision Transformers.}
In response to the success of the transformer architecture in terms of clean accuracy, many works study the performance of ViTs on robustness benchmarks \cite{hendrycks2018benchmarking,hendrycks2021many}. 
It becomes evident that transformers scale better with model size than CNNs and often surpass their convolutional counterparts when faced with corruptions \cite{qin2022understanding,tian2022deeper}, adversarial attacks \cite{lovisotto2022give} %mahmood2021robustness
and distribution shifts \cite{zhang2022delving}. Training \cite{qin2022understanding} and test time \cite{zhang2022memo} augmentation methods utilizing AugMix \cite{hendrycks2019augmix} as well as attention enhancement techniques \cite{chefer2022optimizing} have shown great improvement in robustness. Architectural changes such as position-based attention scaling \cite{mao2022towards} and Fully Attentional Networks (FANs) \cite{zhou2022understanding} further improve performance. Despite these advances, the use of robust transformers in data-constrained domains remains an emerging area of research.

\textbf{Data Augmentation with synthetic images.}
Driven by the data hunger of modern deep learning models, exploiting information from synthetic data to expand training data for downstream tasks is gaining attraction in the field of computer vision. 
In the past, a plethora of works utilized synthetic datasets especially in domains requiring complex annotation such as human motion understanding \cite{pumarola20193dpeople} or semantic segmentation \cite{sankaranarayanan2018learning}.
Early publications focusing on image classification are tailored to specific domains, lacking investigations on general applicability \cite{jaipuria2020deflating}, partly not even deriving performance advantages \cite{bissoto2021gan}.
% Citation options for lacking general applicability: \cite{kumar2022evaluation}

Most general approaches focus on studying the effectiveness of substituting the real training set completely by generated data \cite{ravuri2019classification}, %besnier2020dataset
thereby overlooking the potential of expanding the real dataset with synthetic images. For some approaches, this could be attributed to diminishing performance improvements when combining real and generated data \cite{bissoto2021gan}. %bissoto2021gan he2023is
More recent works employ pre-trained text-to-image models \cite{rombach2022high} %nichol2022glide
to create synthetic classification datasets showing potential of being capable to replace \cite{sariyildiz2023fake,he2023is} and augment \cite{azizi2023synthetic} general object-level datasets. Though, these approaches lack applicability to domains that deviate from common objects, such as medical imaging \cite{medmnistv2}.
% Citation options for text-to-image modesl: \cite{saharia2022photorealistic}
Some works apply generated images to modern transformer models \cite{azizi2023synthetic,he2023is}, others consider the effect of synthetic images on domain shifts and adversarial robustness \cite{gowal2021improving,sariyildiz2023fake}. However, to the best of our knowledge, no research explores the impact of data augmentation with generated images on the robustness of ViTs in limited-data scenarios.

In our work, we explore how generative data augmentation impacts transformer models' robustness in small-scale data scenarios. Demonstrating that augmenting real datasets with images generated by a generative model improves accuracy and robustness against corruptions and domain shifts, our GenFormer approach proves to be versatile across domains. Additionally, we highlight its synergy with conventional methods like augmentation, knowledge distillation, and architectural adaptations, resulting in state-of-the-art performance across models trained from scratch.

% In contrast to downstream-agnostic approaches \cite{sariyildiz2023fake, he2023is} we propose a training strategy applicable regardless of the target domain. 

% The proposed data augmentation strategy exhibits 
% This research avoids the focus on specific-domain applicability and rather proves the concept of utilizing synthetic images to expand small-scale datasets in general. Hence, by examining the performance on general object-level datasets such as CIFAR and Tiny ImageNet this work studies the augmentation effect not only on clean accuracy but also on robustness and distribution shift capability. 
% superiority

\section{Method}
\label{sec:method}

\begin{figure*}[t]
\centering
  \includegraphics[width=.99\linewidth]{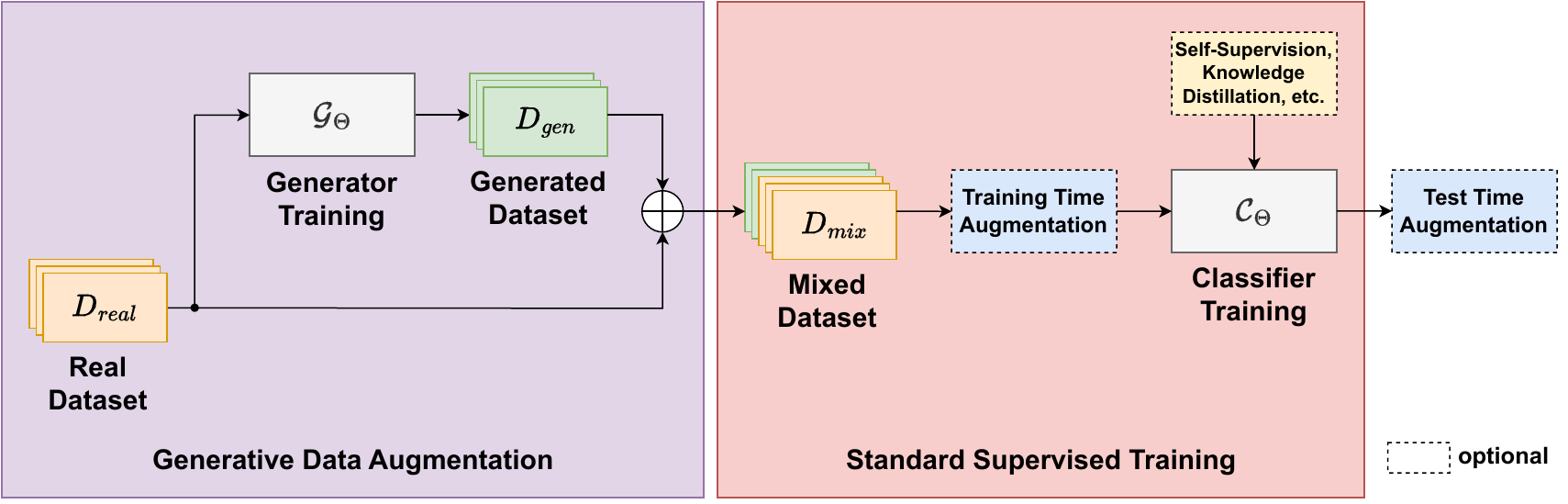}
  \caption{The proposed GenFormer approach involves training a downstream-aware image generation model, $\mathcal{G}_{\Theta}$, using real data $D_{real}$, then augmenting this dataset with generated data $D_{gen}$ to create $D_{mix}$. Subsequently, $\mathcal{C}_{\Theta}$ is trained on $D_{mix}$ for the classification task, with optional methods like data augmentation or knowledge distillation during training. $\oplus$ denotes a concatenation.}
\label{fig:method}
\vspace{-3mm}
\end{figure*}

The aim of this work is to utilize the inherent robustness of the transformer architecture in limited-data scenarios. We therefore propose GenFormer, a generative data augmentation strategy, that alleviates the demand of transformers for large-scale datasets by tackling data scarcity directly, before common training schemes are applied. We accomplish this by exploiting information from generated images, which are created with knowledge about the label space of the downstream task. We refer to this as downstream-aware image generation. The exact procedure can be seen in Figure \ref{fig:method}.

Let $D_{real} = \{x^{real}_i, y^{real}_i\}^{N_{real}}_{i=1}$ be a training dataset where $N_{real}$ denotes the number of real image-label pairs consisting of an image $x^{real}$ and its corresponding class label $y^{real}$. Instead of simply following a standard supervised training strategy solely relying on the real training data, the proposed approach leverages additional information from the real dataset. This is achieved by a generative data augmentation utilizing a class-conditional downstream-aware image generation model $\mathcal{G}_{\Theta}$ learning the parameters $\Theta$ to approximate the true data distribution following the underlying objective $p_{\mathcal{G}}(x|y) \thickapprox p^*(x|y)$. Here, $x$ and $y$ denote a random image with its corresponding label of the underlying data distribution whereas $p_\mathcal{G}$ and $p^*$ denote the learned probability distribution of the generator and the true distribution respectively. The generator is then used to create $N_{gen}$ images $x^{gen} = \mathcal{G}_{\Theta}(z, y_{gen})$ given a random noise vector $z$ and a label $y_{gen}$ as input. This results in an additional generated training dataset $D_{gen} = \{x^{gen}_i, y^{gen}_i\}^{N_{gen}}_{i=1}$. We then combine the real and generated data to receive 
$D_{mix} = \{D_{real}, D_{gen}\}$
which then is used to train a classifier $\mathcal{C}_\Theta$ on the downstream task. In the base setting we follow a standard supervised training.

% fig:real_fake
\begin{figure}[t]
\centering
\includegraphics[width=.99\linewidth]{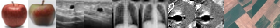} 
\caption{Real (left) and generated (right) sample pairs of corresponding classes (f.l.t.r): CIFAR-100 \cite{krizhevsky2009learning}, BreastMNIST, \mbox{PneumoniaMNIST}, OrganSMNIST \cite{medmnistv2} and EuroSAT \cite{helber2019eurosat}.}
\label{fig:real_fake}
\vspace{-5mm}
\end{figure}

As our GenFormer approach does not depend on a specific generator model, it is compatible with a wide range of image generation models. Based on downstream performance, we employ the EDM \cite{Karras2022edm} diffusion model, which is showcased in Section \ref{sec:generators}.
Since the generator model is trained on the real dataset, the amount of real training images also affects the generator model performance, which may become apparent in lower diversity of the generated images. %, as depicted in Figure \ref{fig:example_less}.
However, in Section \ref{sec:limited} we study the impact of the number of real images available for our method demonstrating consistent improvement through our approach.
A big advantage of the proposed GenFormer approach results from its implementation before the standard downstream training. Thereby, our method allows the additional use of established techniques to increase accuracy and robustness, as can be seen in Figure \ref{fig:method}. In Section \ref{sec:tinyin} we demonstrate the synergy of our approach with conventional data augmentation during train and test time as well as knowledge distillation and architectural modifications.

\section{Evaluation}
\label{sec:eval}

For our evaluation, we conduct extensive experiments involving a range of state-of-the-art Vision Transformers (ViTs) \cite{touvron2021training,wang2021pyramid,wang2022pvt,d2021convit} and convolutional neural networks (CNNs) \cite{he2016deep,woo2023convnext}. 
Unless otherwise mentioned, we employ EDM \cite{Karras2022edm} as generator model $\mathcal{G}_{\Theta}$ to generate additional data, effectively expanding our datasets (see Section \ref{sec:generators}).
Most training runs are carried out with the Tiny ImageNet \cite{le2015tiny} dataset.
For robustness and generalization investigations, we also include our novel Tiny ImageNet versions of various established ImageNet benchmarks, Tiny ImageNetV2, -R, and -A, as well as Tiny ImageNet-C \cite{hendrycks2018benchmarking} in our studies.

In addition to Tiny ImageNet, we extend our investigation to the CIFAR-10 and CIFAR-100 datasets \cite{krizhevsky2009learning}, using CIFAR-10-C and CIFAR-100-C \cite{hendrycks2018benchmarking}. We also consider the test set of CIFAR10.1 \cite{recht2019imagenet}, which contains more challenging samples.
We analyzed three datasets of MedMNIST \cite{medmnistv2} from the medical field, which is known to have limited data. The aim of this study is to extend our evaluation beyond the domain of natural images. In addition, we investigated the impact of our method on the classification of aerial images using EuroSAT \cite{helber2019eurosat}.
Examples of generated data are shown in Figure \ref{fig:real_fake}.
To maintain consistency in our evaluation, we report error rates rather than accuracy, with smaller values indicating better performance.

\subsection{Robustness and Generalization Benchmarks}
The evaluation of the robustness of neural networks plays a major role in real-world applications. 
More specifically, we focus on evaluating different types of robustness of ViTs on small datasets, which has not received much attention in research. To this end, we port the most common robustness and generalization benchmarks \cite{recht2019imagenet,hendrycks2021nae,hendrycks2021many,hendrycks2018benchmarking} to Tiny ImageNet, MedMNIST \cite{medmnistv2} and EuroSAT \cite{helber2019eurosat}.

We create Tiny ImageNetV2 to analyze the generalization ability by keeping all images of joint classes of Tiny ImageNet and ImageNetV2 \cite{recht2019imagenet}.
In the same way, we introduce Tiny ImageNet-R to study the robustness of models when confronted with domain shifts, such as changes in the type of images (e.g. paintings, toys or graffiti). 
Lastly, we propose Tiny ImageNet-A. Based on ImageNet-A \cite{hendrycks2021nae}, we use all images from the original Tiny ImageNet validation set only keeping the images misclassified by ResNet-18 \cite{he2016deep}. 
To further evaluate robustness, we utilize Tiny ImageNet-C of Hendrycks et al. \cite{hendrycks2018benchmarking}, where the validation data from Tiny ImageNet is subjected to 15 different corruptions, each at five severity levels.
Analogous to Tiny ImageNet-C, we introduce novel corrupted test set variants for established image classification benchmarks: EuroSAT-C, which targets aerial imagery, and MedMNIST-C, which focuses on the medical domain. To maintain the integrity of the medical data, we have excluded any weather-dependent corruptions (Snow, Frost, Fog).
For evaluation, we use the Mean Corrupted Error (mCE).
Further details on the test sets can be found in the Appendix (Section \ref{sec:appendix_test_sets}).

\subsection{Comparisons on Tiny ImageNet}
\label{sec:tinyin}

In our first experiment, we perform a comparative analysis of our GenFormer approach in combination with established methods to improve the robustness of neural networks. %, as described in Section \ref{sec:rw}. 
Two transformer-based classifiers, the tiny versions of DeiT \cite{touvron2021training} (without distillation token) and PVT \cite{wang2021pyramid}, are used for this comparison.

As shown in Table \ref{tab:sota}, our experiment demonstrates how our GenFormer approach seamlessly complements various techniques, resulting in notable improvements in both accuracy and robustness. 
As part of our comparative analysis, we employ well-established methods, including CutMix \cite{yun2019cutmix}, Mixup \cite{zhang2018mixup} and AugMix \cite{hendrycks2019augmix} for data augmentation, Locality Guidance \cite{li2022locality} for knowledge distillation, and MEMO \cite{zhang2022memo} for test time augmentation (TTA).
All of these methods are applied on top of our baseline, which is trained for 300 epochs on Tiny ImageNet.
A detailed description of the training can be found in the Appendix (Section \ref{sec:params}).
To expand the dataset with generated images, we use our method to create an additional 100,000 images (corresponds to 100\% of the original dataset).

% tab:sota
\begin{table*}[t]
\centering
\scriptsize
\caption{Analysis of our GenFormer in combination with established SOTA approaches on two Vision Transformers \cite{touvron2021training,wang2021pyramid} on Tiny ImageNet \cite{le2015tiny} and its robustness benchmarks \cite{hendrycks2018benchmarking}. 100,000 generated images are added to the real data.}
\label{tab:sota}
\begin{tabular}{llclllll}
\hline
\multicolumn{1}{c}{\multirow{2}{*}{\textbf{Model}}} & \multicolumn{2}{c|}{\multirow{2}{*}{\textbf{Train-Strategy}}}
& \multicolumn{1}{c}{\textbf{T-IN}}  & \multicolumn{1}{c}{\textbf{T-INv2}} & \multicolumn{1}{c}{\textbf{T-IN-R}} & \multicolumn{1}{c}{\textbf{T-IN-C}} & \multicolumn{1}{c}{\textbf{T-IN-A}}\\
\multicolumn{1}{c}{}  & \multicolumn{2}{c|}{}  & \multicolumn{1}{c}{\textbf{err.}} & \multicolumn{1}{c}{\textbf{err.}}  & \multicolumn{1}{c}{\textbf{err.}} & \multicolumn{1}{c}{\textbf{mCE}} & \multicolumn{1}{c}{\textbf{err.}}  \\ \hline
%%%%%%%%%%%%%%%%%%% Baseline Deit %%%%%%%%%%%%%%%%%%%
\multirow{10}{*}{\textit{DeiT-Ti  \cite{touvron2021training}}}
    & \multirow{2}{*}{Baseline} & \multicolumn{1}{l|}{w/o ours} & 50.3 &  68.0 &  92.5 & 80.6 & 80.6               \\
    & & \multicolumn{1}{l|}{w/ ours} & 44.1 \color{ForestGreen}{(-6.2)} & 65.3 \color{ForestGreen}{(-2.7)}  & 89.6 \color{ForestGreen}{(-2.9)} & 77.7  \color{ForestGreen}{(-2.9)} & 78.0  \color{ForestGreen}{(-2.6)}            \\ \cline{2-8}
    %%%%%%%%%%%%%%%%%%% Cutmix Deit %%%%%%%%%%%%%%%%%%%
    & \multirow{2}{*}{\begin{tabular}[c]{@{}l@{}}CutMix \cite{yun2019cutmix}\\+ Mixup \cite{zhang2018mixup}\end{tabular}} 
    & \multicolumn{1}{l|}{w/o ours} & 44.4   &   65.0   &      89.7   &       74.5   &     78.3         \\
    & & \multicolumn{1}{l|}{w/ ours}  & 38.5  \color{ForestGreen}{(-5.9)} & 58.7  \color{ForestGreen}{(-6.3)} & 86.4   \color{ForestGreen}{(-3.3)} & 71.8  \color{ForestGreen}{(-2.7)} & 74.4  \color{ForestGreen}{(-3.9)}\\ \cline{2-8}
    %%%%%%%%%%%%%%%%%%% AugMix Deit %%%%%%%%%%%%%%%%%%%
    & \multirow{2}{*}{AugMix \cite{hendrycks2019augmix}}                                                       & \multicolumn{1}{l|}{w/o ours} & 40.4         & 61.3          & 88.4          & 72.8          & 76.2         \\
    & & \multicolumn{1}{l|}{w/ ours}  & 36.4 \color{ForestGreen}{(-4.0)} & 57.8 \color{ForestGreen}{(-3.5)} & 85.6 \color{ForestGreen}{(-2.8)} & 71.0 \color{ForestGreen}{(-1.8)} &  73.1 \color{ForestGreen}{(-3.1)} \\ \cline{2-8}
    %%%%%%%%%%%%%%%%%%% Locality Deit %%%%%%%%%%%%%%%%%%%
    & \multirow{2}{*}{\begin{tabular}[c]{@{}l@{}}Locality\\ Guidance \cite{li2022locality}\end{tabular}} & \multicolumn{1}{l|}{w/o ours} & 36.7         & 59.4          & 83.6          & 72.0          & 74.7          \\
    & & \multicolumn{1}{l|}{w/ ours}  & 36.2 \color{ForestGreen}{(-0.5)} & 59.4 ($\pm$0.0) & 84.8 \color{BrickRed}{(+1.2)} & 70.7 \color{ForestGreen}{(-1.3)} & 72.8 \color{ForestGreen}{(-1.3)} \\ \cline{2-8}
    %%%%%%%%%%%%%%%%%%% MEMO Deit %%%%%%%%%%%%%%%%%%%
    & \multirow{2}{*}{MEMO \cite{zhang2022memo}}      & \multicolumn{1}{l|}{w/o ours} & 48.9             & 64.8              &      99.7*         & 78.1              &     99.7*         \\
    & & \multicolumn{1}{l|}{w/ ours}  &  43.1 \color{ForestGreen}{(-5.8)}            &   60.2 \color{ForestGreen}{(-4.6)}            & 99.7*              & 75.5 \color{ForestGreen}{(-3.6)}              & 99.8*              \\ \hline\hline

%%%%%%%%%%%%%%%%%%% Baseline PVT %%%%%%%%%%%%%%%%%%%
\multirow{10}{*}{\textit{PVT-T \cite{wang2021pyramid}}}
    & \multirow{2}{*}{Baseline}                                                       & \multicolumn{1}{l|}{w/o ours} & 46.5 & 67.2  & 91.5   & 78.9   &  78.9   \\
    & & \multicolumn{1}{l|}{w/ ours}  & 42.8 \color{ForestGreen}{(-3.7)} & 64.8 \color{ForestGreen}{(-2.4)} & 87.4 \color{ForestGreen}{(-4.1)} & 76.8 \color{ForestGreen}{(-1.8)} & 77.7 \color{ForestGreen}{(-1.2)} \\  \cline{2-8}
    %%%%%%%%%%%%%%%%%%% Mixup PVT %%%%%%%%%%%%%%%%%%%
    & \multirow{2}{*}{\begin{tabular}[c]{@{}l@{}}CutMix \cite{yun2019cutmix}\\+ Mixup \cite{zhang2018mixup}\end{tabular}}                                                          & \multicolumn{1}{l|}{w/o ours} & 41.9         & 64.6          & 87.8          & 74.1          & 76.8          \\
    & & \multicolumn{1}{l|}{w/ ours}  & 37.5 \color{ForestGreen}{(-4.4)} & 60.1 \color{ForestGreen}{(-4.5)} & 84.6 \color{ForestGreen}{(-3.2)} & 70.1 \color{ForestGreen}{(-4.0)} & 74.1 \color{ForestGreen}{(-2.7)} \\  \cline{2-8}
    %%%%%%%%%%%%%%%%%%% AugMix PVT %%%%%%%%%%%%%%%%%%%
    & \multirow{2}{*}{AugMix \cite{hendrycks2019augmix}} & \multicolumn{1}{l|}{w/o ours} & 39.9 & 62.1 & 87.9 & 73.0 & 75.9 \\
    & & \multicolumn{1}{l|}{w/ ours}  & 36.4 \color{ForestGreen}{(-3.5)} & 58.3 \color{ForestGreen}{(-3.8)} & 86.2 \color{ForestGreen}{(-1.7)} & 71.0 \color{ForestGreen}{(-2.0)} & 73.6 \color{ForestGreen}{(-2.3)}\\  \cline{2-8}
    %%%%%%%%%%%%%%%%%%% Locality PVT %%%%%%%%%%%%%%%%%%%
    & \multirow{2}{*}{\begin{tabular}[c]{@{}l@{}}Locality\\ Guidance \cite{li2022locality}\end{tabular}} & \multicolumn{1}{l|}{w/o ours} & 36.3         & 58.5          & 84.9          & 72.3          & 72.2          \\
    & & \multicolumn{1}{l|}{w/ ours}  & 35.7 \color{ForestGreen}{(-0.6)} & 58.1 \color{ForestGreen}{(-0.4)} & 84.2 \color{ForestGreen}{(-0.7)} & 71.0 \color{ForestGreen}{(-1.3)} & 72.1 \color{ForestGreen}{(-0.1)}              \\  \cline{2-8}
    %%%%%%%%%%%%%%%%%%% MEMO PVT %%%%%%%%%%%%%%%%%%%
   &  \multirow{2}{*}{MEMO \cite{zhang2022memo}}                                                             & \multicolumn{1}{l|}{w/o ours} &  45.1            & 64.5              &  99.8*             &   76.6            & 99.6*              \\
    & & \multicolumn{1}{l|}{w/ ours}  & 42.3 \color{ForestGreen}{(-2.8)}             & 61.4 \color{ForestGreen}{(-3.1)}              & 99.7*              &  75.1 \color{ForestGreen}{(-1.5)}              & 99.6*               \\ \hline
    \multicolumn{8}{l}{* leads to unstable results despite hyperparameter-tuning.}
\end{tabular}
\vspace{-3mm}
\end{table*}

The results in Table \ref{tab:sota} demonstrate the substantial benefits of incorporating additional generated data for both models. 
In the case of DeiT-Ti there is a significant reduction in the base error of $-6.2$, which drops from the original $50.3$ to $44.1$. 
PVT-T initially shows a lower error even without additional data, but still benefits from an additional reduction of $-3.7$ to $42.8$.
Considering all presented robustness benchmarks, the errors of DeiT can be reduced by up to $-2.9$, while the errors of PVT are reduced by up to $-4.1$. 
These outcomes underline the significant potential of leveraging generated data.
Combining our GenFormer approach with various data augmentation techniques such as Mixup and CutMix as commonly used with transformer classifiers and proposed by Touvron et al. \cite{touvron2021training}, generated data is consistently shown to be beneficial for training.
Notably, the combination of generated data with CutMix and Mixup results in a noteworthy improvement of $-5.9$ in Tiny ImageNet validation and an even more substantial reduction of $-6.3$ in Tiny ImageNetV2 for DeiT.

In addition to training time augmentation, we also explore the option of TTA, which is exemplified in our experiments through MEMO. 
When applied to our baseline networks, MEMO yields a moderate improvement, even without the inclusion of generated data. 
However, when we apply TTA to the baseline with generated data, we observe a significant higher enhancement.
Furthermore, despite hyperparameter tuning, MEMO is not performing well on all benchmarks and can have a negative impact on the robustness of the models, especially in the case of Tiny ImageNet-R and -A.
This emphasizes the advantages of generated data, as they are comparatively easy to integrate into the training pipeline and offer significant potential to increase accuracy and robustness.

When including Locality Guidance with a CNN teacher (ResNet-56) trained on the same data as the transformers, our GenFormer approach reduces the errors on clean data for both networks.
However, we observe no improvement for Tiny ImageNetV2 and a slight degradation of $+1.2$ on Tiny ImageNet-R for DeiT.
For all other benchmarks, we see an improvement in results.
The performance gains achieved through generated data are moderate compared to training and test time augmentation, as our approach, like Locality Guidance, contributes to a better focus on more localized features.
This is shown in the mean attention distances comparison presented in the Appendix (Section \ref{sec:attention_maps}).

\subsection{Limited-Data Analysis on Tiny ImageNet}
\label{sec:limited}

\begin{figure*}[t]
\centering
  \includegraphics[width=.95\linewidth]{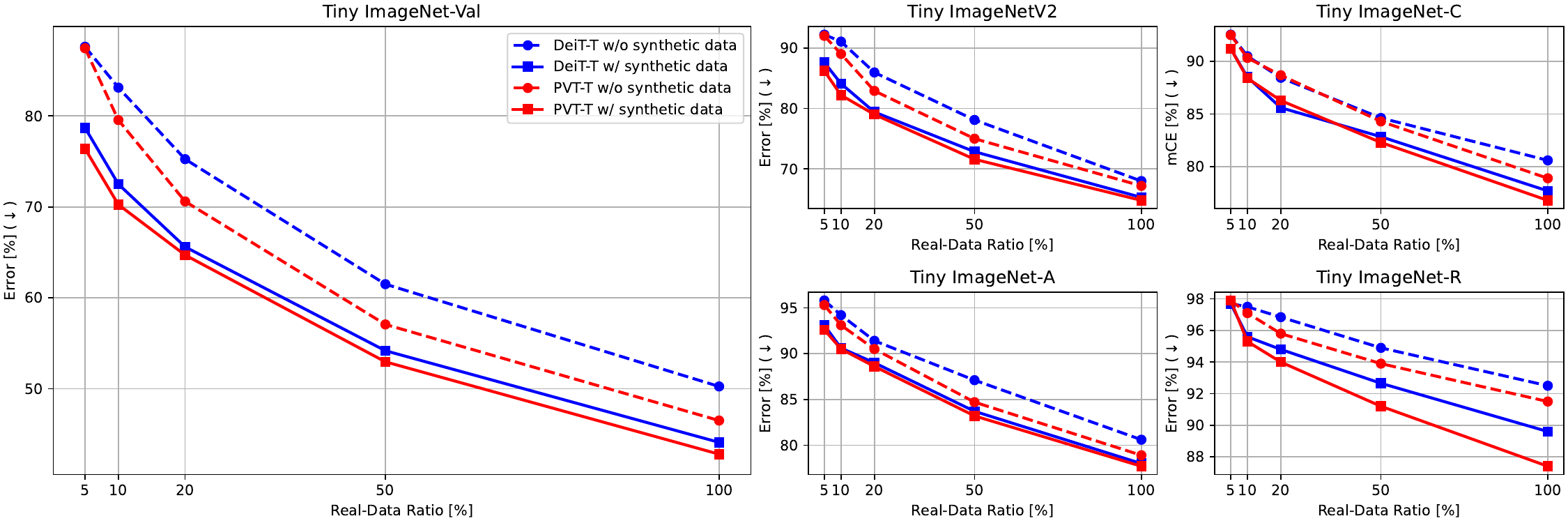}  
  \caption{Analysis of the influence of different amounts of training data on the accuracy and robustness of Vision Transformers \cite{touvron2021training,wang2021pyramid}. The networks are trained with \{5, 10, 20, 50, 100\}\% of Tiny ImageNet \cite{le2015tiny}. We add 100,000 generated images (from a diffusion model trained on the same amount) to each train-set.}
  \label{fig:subset}
  \vspace{-3mm}
\end{figure*}

In our second experiment, we investigate the impact of limited data on the training and robustness of ViTs and the diffusion model used to generate the data. For this purpose, we divide the train set of Tiny ImageNet into four random subsets of 5\%, 10\%, 20\% and 50\%. 
In addition, we also run the experiment on the full set of data. 
In the first step, we trained the diffusion model for each of these subsets and generated a total of 100,000 images. 
In the second step, we carried out training sessions for PVT-T and DeiT-Ti, both with and without the inclusion of generated data. 
The results, including error rates on the validation set and robustness benchmarks, are presented in Figure \ref{fig:subset}.

Congruent with previous findings, the accuracy of ViTs increases with the amount of real data.
For instance, the error rate of DeiT, which starts at $87.6$ when no generated images are added to a subset of 5\%, decreases to $75.2$ with 20\% of the real data, and further drops to $50.3$ when all 100,000 real images are utilized for training.
Robustness across all benchmarks follows the same trend, showing consistent and steady improvement. 
When training with our method, there is a substantial reduction of the error and an increase of robustness, for both PVT and DeiT. 
Interestingly, even models with only 5\% real data in conjunction with generated data achieve higher accuracy than models trained with 10\% real images. 
This pattern is also evident when comparing models with 10\% real data paired with generated images to those with 20\% real data but without generated data.
%Our method effectively prevents overfitting as shown in Figure \ref{fig:overfit}. In this example, DeiT-Ti trained on the 20\% subset strongly tends to overfit, as can be seen from the test loss (blue line). This problem is mitigated by including the generated data, as illustrated by the red line.
Nevertheless, using a small amount of real data (e.g. 5\%, 10\%, or 20\%) in conjunction with a large amount of generated data is not as effective as training with a larger amount of real data (e.g. 50\% or 100\%). 

\subsection{Going beyond Natural Images} 
\label{sec:domain}
\begin{table*}[t]
\centering
\caption{Analysis of GenFormer for training multiple Vision Transformer \cite{touvron2021training,wang2021pyramid,wang2022pvt,d2021convit,mao2022towards,zhou2022understanding} and  Convolutional Neural Networks \cite{he2016deep,woo2023convnext} on three datasets of the MedMNIST \cite{medmnistv2,pneumoniamnist,organmnist1,breastmnist} collection (medical domain) and on EuroSAT\cite{helber2019eurosat} (aerial domain) as well as their corresponding robustness benchmarks \cite{hendrycks2018benchmarking}. }
\label{tab:domain}
\tiny
\centering
\scalebox{0.97}{
\begin{tabular}{llllllllll}
\hline
\multicolumn{2}{c|}{\multirow{2}{*}{\textbf{Model}}} & \multicolumn{2}{c|}{\textbf{B-MNIST(-C)} } & \multicolumn{2}{c|}{\textbf{P-MNIST(-C)} } & \multicolumn{2}{c|}{\textbf{OS-MNIST(-C)} } & \multicolumn{2}{c}{\textbf{EuroSAT(-C)} } \\
\multicolumn{2}{c|}{} & \multicolumn{1}{c}{\textbf{err.}} & \multicolumn{1}{c|}{\textbf{mCE}} & \multicolumn{1}{c}{\textbf{err.}} & \multicolumn{1}{c|}{\textbf{mCE}} & \multicolumn{1}{c}{\textbf{err.}} & \multicolumn{1}{c|}{\textbf{mCE}} & \multicolumn{1}{c}{\textbf{err.}} & \multicolumn{1}{c}{\textbf{mCE}} \\ \hline

\multicolumn{10}{c}{\textit{Convolutional Neural Networks}} \\ \hline

%%%%%%%%%%%%%%%%%%% ResNet-18 %%%%%%%%%%%%%%%%%%%
\multirow{2}{*}{ResNet-18 \cite{he2016deep}} & \multicolumn{1}{l|}{w/o ours} & 17.9 & \multicolumn{1}{l|}{19.8} & 4.2 & \multicolumn{1}{l|}{8.6} & 17.4 & \multicolumn{1}{l|}{30.4} & 1.5 & 24.6 \\
& \multicolumn{1}{l|}{w/ ours} & 8.3 \color{ForestGreen}{(-9.6)} & \multicolumn{1}{l|}{15.6 \color{ForestGreen}{(-4.2)}} & 3.7 \color{ForestGreen}{(-0.5)} & \multicolumn{1}{l|}{7.7 \color{ForestGreen}{(-0.9)}} & 16.8 \color{ForestGreen}{(-0.6)} & \multicolumn{1}{l|}{26.5 \color{ForestGreen}{(-3.9)}} & 1.1 \color{ForestGreen}{(-0.4)} & 23.1 \color{ForestGreen}{(-1.5)} \\

%%%%%%%%%%%%%%%%%%% ConvNeXtv2-F %%%%%%%%%%%%%%%%%%%
Conv- & \multicolumn{1}{l|}{w/o ours} & 19.2 & \multicolumn{1}{l|}{21.5} & 4.5 & \multicolumn{1}{l|}{7.0} & 17.5 & \multicolumn{1}{l|}{25.7} & 1.4 & 22.7 \\
NeXtv2-F \cite{woo2023convnext} & \multicolumn{1}{l|}{w/ ours} & 9.0 \color{ForestGreen}{(-10.2)} & \multicolumn{1}{l|}{12.9 \color{ForestGreen}{(-8.6)}} & 4.2 \color{ForestGreen}{(-0.3)} & \multicolumn{1}{l|}{7.7 \color{BrickRed}{(+0.7)}} & 16.2 \color{ForestGreen}{(-1.3)} & \multicolumn{1}{l|}{23.2 \color{ForestGreen}{(-2.5)}} & 0.9 \color{ForestGreen}{(-0.5)} & 21.9 \color{ForestGreen}{(-0.8)} \\ \hline

\multicolumn{10}{c}{\textit{Pure Vision Transformer}} \\ \hline

%%%%%%%%%%%%%%%%%%% DeiT-Ti %%%%%%%%%%%%%%%%%%%
\multirow{2}{*}{DeiT-Ti \cite{touvron2021training}} & \multicolumn{1}{l|}{w/o ours} & 23.1 & \multicolumn{1}{l|}{24.5} & 8.0 & \multicolumn{1}{l|}{12.8} & 26.9 & \multicolumn{1}{l|}{45.9} & 2.4 & 27.7 \\
& \multicolumn{1}{l|}{w/ ours} & 9.0 \color{ForestGreen}{(-14.1)} & \multicolumn{1}{l|}{14.7 \color{ForestGreen}{(-9.8)}} & 5.1 \color{ForestGreen}{(-2.9)} & \multicolumn{1}{l|}{10.5 \color{ForestGreen}{(-2.3)}} & 21.3 \color{ForestGreen}{(-5.6)} & \multicolumn{1}{l|}{32.5 \color{ForestGreen}{(-13.4)}} & 1.7 \color{ForestGreen}{(-0.7)} & 26.9 \color{ForestGreen}{(-0.8)} \\

%%%%%%%%%%%%%%%%%%% PVT-T %%%%%%%%%%%%%%%%%%%
\multirow{2}{*}{PVT-T \cite{wang2021pyramid}} & \multicolumn{1}{l|}{w/o ours} & 22.4 & \multicolumn{1}{l|}{23.9} & 13.1 & \multicolumn{1}{l|}{17.8} & 20.3 & \multicolumn{1}{l|}{34.4} & 1.9 & 26.2 \\
& \multicolumn{1}{l|}{w/ ours} & 9.0 \color{ForestGreen}{(-13.4)} & \multicolumn{1}{l|}{13.9 \color{ForestGreen}{(-10.0)}} & 5.4 \color{ForestGreen}{(-7.7)} & \multicolumn{1}{l|}{10.0 \color{ForestGreen}{(-7.8)}} & 17.1 \color{ForestGreen}{(-3.2)} & \multicolumn{1}{l|}{24.7 \color{ForestGreen}{(-9.7)}} & 1.2 \color{ForestGreen}{(-0.7)} & 25.4 \color{ForestGreen}{(-0.8)} \\ \hline
    
\multicolumn{10}{c}{\textit{Hybrid Architectures}} \\ \hline

%%%%%%%%%%%%%%%%%%% PVTv2-B0 %%%%%%%%%%%%%%%%%%%
\multirow{2}{*}{PVTv2-B0 \cite{wang2022pvt}} & \multicolumn{1}{l|}{w/o ours} & 16.7 & \multicolumn{1}{l|}{18.5} & 5.9 & \multicolumn{1}{l|}{9.1} & 17.0 & \multicolumn{1}{l|}{28.4} & 1.5 & 24.0 \\
& \multicolumn{1}{l|}{w/ ours} & 9.0 \color{ForestGreen}{(-7.7)} & \multicolumn{1}{l|}{13.3 \color{ForestGreen}{(-5.2)}} & 4.5 \color{ForestGreen}{(-1.4)} & \multicolumn{1}{l|}{8.9 \color{ForestGreen}{(-0.2)}} & 16.2 \color{ForestGreen}{(-0.8)} & \multicolumn{1}{l|}{22.6 \color{ForestGreen}{(-5.8)}} & 1.0 \color{ForestGreen}{(-0.5)} & 22.9 \color{ForestGreen}{(-1.1)} \\

%%%%%%%%%%%%%%%%%%% ConViT-Ti %%%%%%%%%%%%%%%%%%%
\multirow{2}{*}{ConViT-Ti \cite{d2021convit}} & \multicolumn{1}{l|}{w/o ours} & 16.0 & \multicolumn{1}{l|}{22.2} & 6.3 & \multicolumn{1}{l|}{11.8} & 19.6 & \multicolumn{1}{l|}{35.3} & 2.6 & 28.0 \\
& \multicolumn{1}{l|}{w/ ours} & 7.7 \color{ForestGreen}{(-8.3)} & \multicolumn{1}{l|}{12.0 \color{ForestGreen}{(-10.2)}} & 3.5 \color{ForestGreen}{(-2.8)} & \multicolumn{1}{l|}{6.8 \color{ForestGreen}{(-5.0)}} & 15.5 \color{ForestGreen}{(-4.1)} & \multicolumn{1}{l|}{22.4 \color{ForestGreen}{(-12.9)}} & 1.4 \color{ForestGreen}{(-1.2)} & 25.7 \color{ForestGreen}{(-2.3)} \\ \hline

\multicolumn{10}{c}{\textit{Robust Architectures}} \\ \hline

%%%%%%%%%%%%%%%%%%% RVT-Ti %%%%%%%%%%%%%%%%%%%
\multirow{2}{*}{RVT-Ti \cite{mao2022towards}} & \multicolumn{1}{l|}{w/o ours} & 10.9 & \multicolumn{1}{l|}{16.6} & 3.2 & \multicolumn{1}{l|}{7.8} & 15.7 & \multicolumn{1}{l|}{24.0} & 1.3 & 22.3 \\
& \multicolumn{1}{l|}{w/ ours} & 7.7 \color{ForestGreen}{(-3.2)} & \multicolumn{1}{l|}{12.2 \color{ForestGreen}{(-4.4)}} & 3.7 \color{BrickRed}{(+0.5)} & \multicolumn{1}{l|}{9.3 \color{BrickRed}{(+1.5)}} & 16.0 \color{BrickRed}{(+0.3)} & \multicolumn{1}{l|}{21.7 \color{ForestGreen}{(-2.3)}} & 1.0 \color{ForestGreen}{(-0.3)} & 21.8 \color{ForestGreen}{(-0.5)} \\

%%%%%%%%%%%%%%%%%%% FAN-T %%%%%%%%%%%%%%%%%%%
\multirow{2}{*}{FAN-T \cite{zhou2022understanding}} & \multicolumn{1}{l|}{w/o ours} & 12.8 & \multicolumn{1}{l|}{16.5} & 4.6 & \multicolumn{1}{l|}{8.1} & 16.0 & \multicolumn{1}{l|}{25.1} & 1.5 & 22.6 \\
& \multicolumn{1}{l|}{w/ ours} & 8.3 \color{ForestGreen}{(-4.5)} & \multicolumn{1}{l|}{12.1 \color{ForestGreen}{(-4.4)}} & 3.4 \color{ForestGreen}{(-1.2)} & \multicolumn{1}{l|}{7.7 \color{ForestGreen}{(-0.4)}} & 15.2 \color{ForestGreen}{(-0.8)} & \multicolumn{1}{l|}{21.8 \color{ForestGreen}{(-3.3)}} & 1.2 \color{ForestGreen}{(-0.3)} & 21.8 \color{ForestGreen}{(-0.8)} \\ \hline 
\end{tabular}}
\vspace{-3mm}
\end{table*}

In contrast to Tiny ImageNet and CIFAR, which consist mainly of natural images, other domains such as medical imaging suffer from a lack of data. It is not unusual for medical datasets to comprise only a few hundred to a few thousand images. Therefore, we assessed the MedMNIST \cite{medmnistv2} collection, which includes PneumoniaMNIST \cite{pneumoniamnist} with 5,856 images, OrganSMNIST \cite{organmnist1} with 25,211 images, and BreastMNIST \cite{breastmnist} with only 780 images. Additionally, we use EuroSAT \cite{helber2019eurosat} to classify aerial images, providing a more comprehensive evaluation across various fine-grained domains.
In order to evaluate the robustness against common corruptions, the proposed MedMNIST-C and EuroSAT-C test sets are utilized.

In addition to pure ViTs such as DeiT-Ti  and PVT-T, we include hybrid architectures \cite{d2021convit,wang2022pvt} combining self-attention with convolutional operations. 
We also investigate pure CNN architectures such as ConvNeXtv2-F \cite{woo2023convnext} and ResNet-18 \cite{he2016deep}. 
Furthermore, we analyze transformer architectures such as RVT-Ti \cite{mao2022towards} and FAN-T \cite{zhou2022understanding}, which are specifically designed to improve robustness.
All models are trained in the same way for 300 epochs on the respective datasets and, if specified, extended by 50k generated images. 
The results are listed in Table \ref{tab:domain}.

Starting with BreastMNIST, the smallest dataset in our analysis, we observe a significant improvement in terms of error on clean data as well as mCE on corrupted data. The pure ViTs benefit the most from the additional data, lowering the error up to $-14.1$ on clean data and $-10.0$ on corrupted data. The CNNs and other transformer architectures also benefit from the additional artificial data, with the robust architectures showing the smallest improvement of about $-4$. A consistent improvement is also observed for the other two medical datasets, with few exceptions. Similar to BreastMNIST, the pure transformers demonstrate the highest gain in performance as a result of additional generated images. When examining the result on EuroSAT, it can be seen that generated data can lead to further improvement even for already low error rates.

\subsection{Comparisons on CIFAR-10 and CIFAR-100}
\label{sec:cifar}

In our last experiment, we extend our investigations for natural benchmarks beyond Tiny ImageNet to the smaller CIFAR-10 and CIFAR-100 datasets. 
In addition, we utilize CIFAR-10.1, CIFAR-10-C, and CIFAR-100-C. 
The purpose of this evaluation is to show the versatility of our GenFormer approach and its positive impact on a variety of architectural models.
For all experiments on CIFAR, we use the same architectures as in Section \ref{sec:domain}.
All models are trained in the same way for 300 epochs on the respective dataset and, if specified, extended by 100,000 generated images. 
The results are listed in Table \ref{tab:cifar}.

\begin{table}[t]
\setlength{\tabcolsep}{1.6mm}
\caption{Analysis of GenFormer for training multiple Vision Transformers with \cite{touvron2021training,wang2021pyramid,wang2022pvt,d2021convit,mao2022towards,zhou2022understanding} and Convolutional Neural Networks \cite{he2016deep,woo2023convnext} on CIFAR \cite{krizhevsky2009learning} and its corresponding robustness benchmarks \cite{hendrycks2018benchmarking,recht2019imagenet}. }
\label{tab:cifar}
\tiny
\centering
\begin{tabular}{llllllll}
\hline
% \multicolumn{2}{c|}{\multirow{2}{*}{\textbf{}}}                                                                     & \multicolumn{3}{c|}{\textbf{CIFAR-10}}              & \multicolumn{2}{c}{\textbf{CIFAR-100}}               \\
\multicolumn{2}{c|}{\multirow{2}{*}{\textbf{Model}}}                                                                     & \multicolumn{1}{c}{\textbf{C-10}} & \multicolumn{1}{c}{\textbf{C-10.1}} & \multicolumn{1}{c|}{\textbf{C-10-C}} & \multicolumn{1}{c}{\textbf{C-100}} & \multicolumn{1}{c}{\textbf{C-100-C}} \\
\multicolumn{2}{c|}{}                                                                                               & \multicolumn{1}{c}{\textbf{err.}} & \multicolumn{1}{c}{\textbf{err.}} & \multicolumn{1}{c|}{\textbf{mCE}} & \multicolumn{1}{c}{\textbf{err.}} & \multicolumn{1}{c}{\textbf{mCE}}  \\ \hline

\multicolumn{7}{c}{\textit{Convolutional Neural Networks}}\\ \hline
\multirow{2}{*}{ResNet-18 \cite{he2016deep}}  & \multicolumn{1}{l|}{w/o ours} & 4.6  & 11.2 & \multicolumn{1}{l|}{15.1}    & 20.5  & 37.9              \\
& \multicolumn{1}{l|}{w/ ours}  & 4.1 \color{ForestGreen}{(-0.5)} & 10.1 \color{ForestGreen}{(-1.1)} & \multicolumn{1}{l|}{14.6 \color{ForestGreen}{(-0.5)}}    & 20.3 \color{ForestGreen}{(-0.2)}  & 38.0 \color{BrickRed}{(+0.1)} \\
\multirow{2}{*}{ConvNeXtv2-F \cite{woo2023convnext}} & \multicolumn{1}{l|}{w/o ours} & 4.3  & 11.0 & \multicolumn{1}{l|}{11.1}  & 24.2   & 38.9  \\
     & \multicolumn{1}{l|}{w/ ours}  & 3.1 \color{ForestGreen}{(-1.2)}  & 7.9 \color{ForestGreen}{(-3.1)} &  \multicolumn{1}{l|}{10.0 \color{ForestGreen}{(-1.1)}}       & 19.1 \color{ForestGreen}{(-5.1)}   & 33.2 \color{ForestGreen}{(-5.7)}          \\ \hline
     
\multicolumn{7}{c}{\textit{Pure Vision Transformer}} \\ \hline
\multirow{2}{*}{DeiT-Ti  \cite{touvron2021training}}  & \multicolumn{1}{l|}{w/o ours} & 10.5   & 22.2 & \multicolumn{1}{l|}{23.0}     & 35.3     & 51.8   \\
    & \multicolumn{1}{l|}{w/ ours}  & 4.0 \color{ForestGreen}{(-6.5)} & 9.8 \color{ForestGreen}{(-12.4)} & \multicolumn{1}{l|}{12.0 \color{ForestGreen}{(-11.0)}}    & 24.7 \color{ForestGreen}{(-10.6)}    & 39.4 \color{ForestGreen}{(-12.4)}      \\
\multirow{2}{*}{PVT-T \cite{wang2021pyramid}}    & \multicolumn{1}{l|}{w/o ours} & 6.9 & 14.8 & \multicolumn{1}{l|}{17.8}    & 29.7  & 48.0  \\
    & \multicolumn{1}{l|}{w/ ours}  &  3.6 \color{ForestGreen}{(-3.3)} & 9.7 \color{ForestGreen}{(-5.1)} & \multicolumn{1}{l|}{11.7 \color{ForestGreen}{(-6.1)}}  & 21.1 \color{ForestGreen}{(-8.6)}   & 35.9 \color{ForestGreen}{(-12.1)}   \\ \hline
    
\multicolumn{7}{c}{\textit{Hybrid Architectures}} \\ \hline
\multirow{2}{*}{PVTv2-B0 \cite{wang2022pvt}}   & \multicolumn{1}{l|}{w/o ours} & 5.0 & 11.1 & \multicolumn{1}{l|}{14.2}  & 23.1  & 41.2 \\
    & \multicolumn{1}{l|}{w/ ours}  & 3.5 \color{ForestGreen}{(-1.5)}  & 8.8 \color{ForestGreen}{(-2.3)} & \multicolumn{1}{l|}{12.6 \color{ForestGreen}{(-1.6)}} & 19.5 \color{ForestGreen}{(-3.6)} & 34.2 \color{ForestGreen}{(-7.0)} \\
\multirow{2}{*}{ConViT-Ti \cite{d2021convit}} & \multicolumn{1}{l|}{w/o ours} & 5.6 & 13.2 &  \multicolumn{1}{l|}{14.0}  & 25.5 & 40.6  \\
    & \multicolumn{1}{l|}{w/ ours}  &  3.2 \color{ForestGreen}{(-2.4)}  & 7.8 \color{ForestGreen}{(-5.4)} & \multicolumn{1}{l|}{9.6 \color{ForestGreen}{(-4.4)}} & 18.2 \color{ForestGreen}{(-7.3)}      & 31.5 \color{ForestGreen}{(-9.1)}    \\ \hline

\multicolumn{7}{c}{\textit{Robust Architectures}} \\ \hline
\multirow{2}{*}{RVT-Ti \cite{mao2022towards}}   & \multicolumn{1}{l|}{w/o ours} & 2.9 & 8.0 & \multicolumn{1}{l|}{9.0}  & 18.1  & 31.1 \\
    & \multicolumn{1}{l|}{w/ ours}  & 2.4 \color{ForestGreen}{(-0.5)}  & 5.8 \color{ForestGreen}{(-2.2)} & \multicolumn{1}{l|}{7.2 \color{ForestGreen}{(-1.8)}} & 15.3 \color{ForestGreen}{(-2.8)} & 27.3 \color{ForestGreen}{(-3.8)} \\
\multirow{2}{*}{FAN-T \cite{zhou2022understanding}} & \multicolumn{1}{l|}{w/o ours} & 3.4 & 9.1 & \multicolumn{1}{l|}{10.0} & 19.8 & 34.3 \\
    & \multicolumn{1}{l|}{w/ ours}  & 2.8 \color{ForestGreen}{(-0.6)} & 8.0 \color{ForestGreen}{(-1.1)} & \multicolumn{1}{l|}{9.1 \color{ForestGreen}{(-0.9)}} & 18.2 \color{ForestGreen}{(-1.6)} & 31.7 \color{ForestGreen}{(-2.6)}\\ \hline
    
\end{tabular}
\vspace{-3mm}
\end{table}

The evaluation results show that pure ViTs benefit the most from the additional data.
The error of DeiT-Ti on CIFAR-100 is reduced by more than $30\%$ relatively compared to the error without additional training data. 
The mCE on the corrupted data is also reduced by about $24\%$. 
In the case of PVT-T, there is a relative improvement of almost $30\%$ and $25\%$, respectively, on the corrupted data.
Significant improvements are also seen in the case of CIFAR-10. 
The hybrid and robust architectures also show improvements. These are in the relative range of about $10$ to $30\%$. 
It should be noted that ResNet-18 as a CNN is almost unaffected by the additional generated data.
Neither the error, nor the robustness improves significantly. 
In contrast to ResNet, ConvNeXtv2-F, which adapts many of the transformer design decisions for CNNs, responds much better to the generated data. 
There is a relative improvement of about $20\%$ over CIFAR-100 and $15\%$ over CIFAR-100-C.
Again, the experiment on CIFAR exhibits the potential of using additional generated data to improve both the accuracy and robustness of especially ViTs.
It demonstrates that the additional generated data helps transformers to learn more local features (see Appendix, Section \ref{sec:attention_maps}). 
Since CNNs naturally inherit a local bias, the gain from additional data diminishes for these networks. Furthermore, our GenFormer approach shows that for small datasets we are able to close the gap between CNNs and ViTs.
In addition to experiments on small models ($<$ 15M parameters), we present further results on CIFAR-100, demonstrating scalability with models of up to 90M parameters and achieving state-of-the-art accuracy and robustness by combining GenFormer with established  methods (included in the Appendix, Section \ref{sec:appendix_big_models}).

\subsection{Ablation Study}
In the subsequent ablation studies, we aim to investigate the influence of both the data generation network and the quality of the generated data on the accuracy and robustness of transformer classifiers. 
In addition, we investigate the effects of a longer training duration compared to a larger dataset.

\subsubsection{Comparison of different Generative Models}
\label{sec:generators}

\begin{table}[t]
\caption{Comparison of Generative Models \cite{sauer2022stylegan,Karras2020ada,kang2021rebooting,brock2018large,kim2022refining,xu2023pfgm++,Karras2022edm} for generating 100,000 images to train a ViT \cite{touvron2021training} on CIFAR-10 \cite{krizhevsky2009learning}. 
%We generated 100,000 images, which were used in two ways: pre-training the network followed by fine-tuning on real data (\textit{Gen. + FT}), or mixing with real data (\textit{Mix}). 
}
\label{tab:ablation}
\footnotesize
\centering
\scalebox{0.85}{
\begin{tabular}{lccccc}
\hline
\multicolumn{1}{c|}{\multirow{2}{*}{\textbf{Generative Models}}} & \multicolumn{1}{c|}{\multirow{2}{*}{\textbf{FID\textsuperscript{*}} $\downarrow$}} & \multicolumn{2}{c|}{\textbf{Gen. + FT}}           & \multicolumn{2}{c}{\textbf{Mix}} \\
\multicolumn{1}{c|}{}                                 & \multicolumn{1}{c|}{}                              & \textbf{err.} & \multicolumn{1}{c|}{\textbf{mCE}} & \textbf{err.}    & \textbf{mCE}   \\ \hline
\multicolumn{1}{l|}{Baseline (w/o gen. data)}                         & \multicolumn{1}{c|}{NA}                            &  10.5            & \multicolumn{1}{c|}{23.0}             &   10.5              &  23.0              \\ \hline
\multicolumn{6}{c}{\textit{Generative Adversarial Networks}} \\ \hline
\multicolumn{1}{l|}{StyleGAN-XL \cite{sauer2022stylegan}}                       & \multicolumn{1}{c|}{1.85}                          &  10.1          & \multicolumn{1}{c|}{21.1}             & 6.5             &  17.1             \\
\multicolumn{1}{l|}{StyleGAN2-ADA \cite{Karras2020ada}}                    & \multicolumn{1}{c|}{2.42}                          &  8.2            & \multicolumn{1}{c|}{19.7}             &  6.4               & 15.4               \\
\multicolumn{1}{l|}{ReACGAN \cite{kang2021rebooting}}                           & \multicolumn{1}{c|}{3.87}                         &   8.1           & \multicolumn{1}{c|}{19.3}             & 6.0                &     15.5           \\
\multicolumn{1}{l|}{StyleGAN2 \cite{Karras2020ada}}                        & \multicolumn{1}{c|}{6.96}                         &  7.8            & \multicolumn{1}{c|}{19.3}             &   6.0              &  15.1              \\
\multicolumn{1}{l|}{BigGAN \cite{brock2018large}}                           & \multicolumn{1}{c|}{14.73}                         &   7.5         & \multicolumn{1}{c|}{18.8}             &  5.6               &   14.8             \\
\hline
\multicolumn{6}{c}{\textit{{Diffusion Models}}}\\ \hline
\multicolumn{1}{l|}{EDM-G++ \cite{kim2022refining}}                           & \multicolumn{1}{c|}{1.64}                              &   6.4            & \multicolumn{1}{c|}{17.2}             &   4.0              & 12.1               \\
\multicolumn{1}{l|}{PFGM++ \cite{xu2023pfgm++}}                           & \multicolumn{1}{c|}{1.74}                          &     6.5         & \multicolumn{1}{c|}{17.4}             &     4.1            &   12.5             \\
\multicolumn{1}{l|}{EDM \cite{Karras2022edm}}                           & \multicolumn{1}{c|}{1.79}                              &   6.8           & \multicolumn{1}{c|}{17.5}             & 4.0                &   12.0             \\
\hline
\multicolumn{6}{l}{* FIDs are from the original publications.} 
\end{tabular}}
\vspace{-5mm}
\end{table}

To begin, our initial focus revolves around assessing the impact of the data generation network on both the accuracy and robustness of transformer networks. 
To achieve this, we conduct a comparative analysis involving state-of-the-art Generative Adversarial Networks (GANs) \cite{sauer2022stylegan,Karras2020ada,brock2018large,kang2021rebooting} and pixel-diffusion models \cite{xu2023pfgm++,Karras2022edm,kim2022refining}. 
We employ CIFAR-10 and ensure fairness by utilizing the best performing weights provided by the respective method's developers.
Since we focus on low-resolution datasets ($32\times32$ for CIFAR and $64\times64$ for Tiny ImageNet \cite{le2015tiny}), our investigation is limited to pixel-diffusion models. 
%It is worth noting that when considering higher resolutions, latent diffusion models \cite{rombach2022high} should also be taken into account.
For this comparative analysis between the mentioned models, we employ DeiT-Ti  as the classifier subjected to two different training approaches. 
In the first scenario, the network undergoes an initial pre-training phase of 200 epochs on 100,000 generated images from the generative models, followed by a fine-tuning of 100 epochs on the real data. 
In the second scenario, we combine real and generated data and perform a single training phase of 300 epochs. 
Throughout the training process, we follow the training strategy provided by Liu et al. \cite{liu2021efficient}.
The results are listed in Table \ref{tab:ablation}.

Our results show that simply comparing FID scores is not sufficient when generating additional data to train ViTs. These findings further support observations of an earlier study by Ravuri et al. \cite{ravuri2019classification}.
In the case of GANs, there is no discernible correlation between FID scores and accuracy. 
In both training variants tested, StyleGAN-XL \cite{sauer2022stylegan}, despite having the best FID ($1.85$), leads to the highest error on clean data ($10.1$ / $6.5$) and on corrupted data ($21.1$ / $17.1$). Conversely, BigGAN \cite{brock2018large}, which has the highest FID, leads to the lowest error ($7.5$ / $5.6$). 
Nevertheless, the use of generated data consistently outperforms the baseline in training, regardless of the quality of the generated data.

In contrast to GANs, we observe significantly lower variability between FID values of diffusion models and the achieved error rates of the transformer network on CIFAR-10. 
Notably, there are no discernible performance disparities among the diffusion models. 
It is noteworthy that all diffusion models consistently outperform the compared GANs, leading to significant improvements in the classification network.
For instance, the error of DeiT-Ti on clean data is reduced from $10.5$ to $4.0$, and on corrupted data, the mCE drops from $23.0$ to $12.0$. 
Given the lack of significant differences between the diffusion models, we consider the results of mixing real data with generated data when choosing our generator model. 
Considering the results in Table \ref{tab:ablation}, we choose EDM \cite{Karras2022edm} due to its competitive generation results and consequently use it alongside the mixing strategy for all our experiments.
\vspace{-3mm}

\subsubsection{Iterations vs. Samples}
\begin{figure}[t]
\centering
\includegraphics[width=0.99\linewidth]{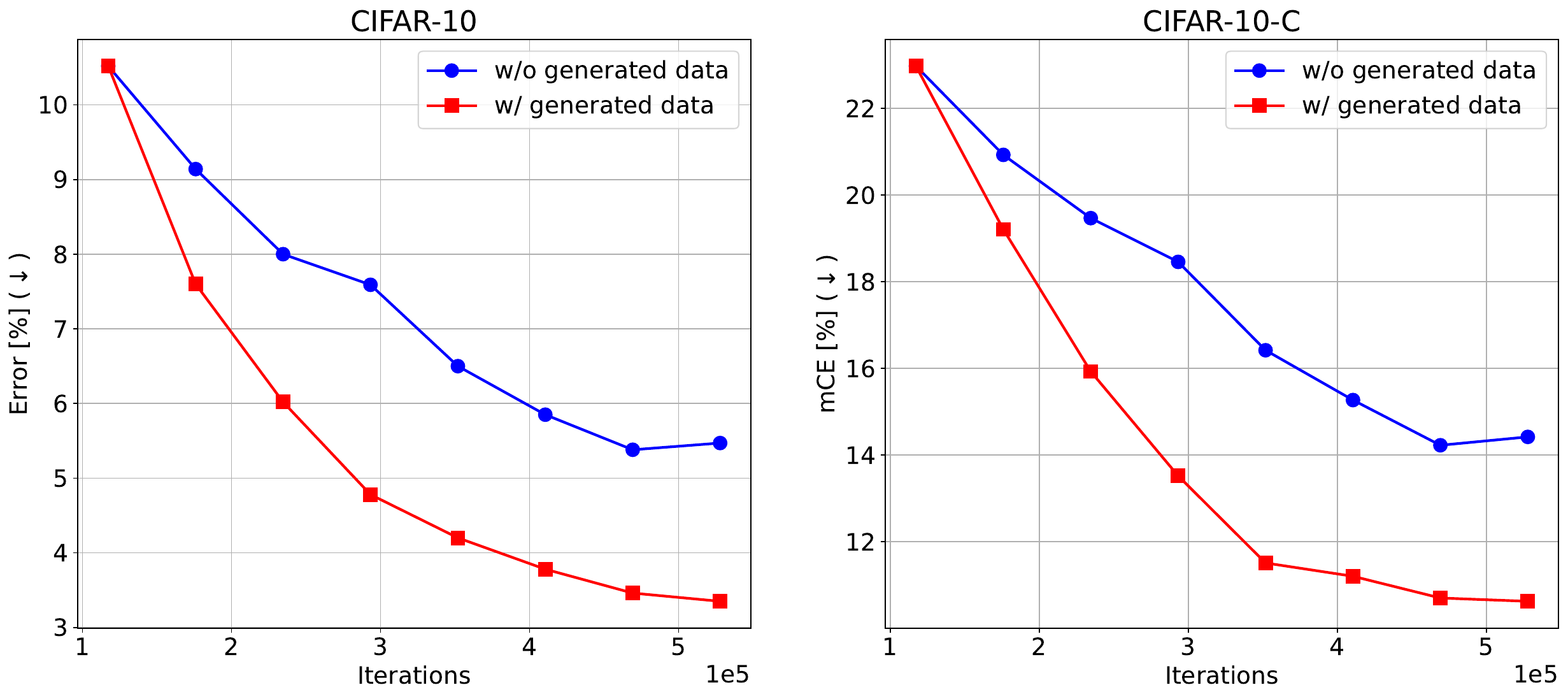} 
\caption{Analysis of the impact of the duration of training (\textcolor{blue}{blue line}) versus the number of data (\textcolor{red}{red line}) of DeiT-Ti  \cite{touvron2021training} on CIFAR-10(-C)\cite{krizhevsky2009learning,hendrycks2018benchmarking}.}
\label{fig:duration}
\vspace{-3mm}
\end{figure}

To validate that the presented improvements are a result of the proposed GenFormer approach and not by a longer training duration, the impact of the amount of training iterations is investigated. 
Figure \ref{fig:duration} shows the comparison of training runs with and without generated images of DeiT-Ti  on CIFAR-10. 
The amount of iterations of the ViT training with generated images is controlled via the amount of additional generated images, whereas the training iterations for the training without generated images is controlled by the number of epochs. 
The graph starts at approximately $1.2 \cdot 10^5$ iterations, which corresponds to 300 epochs. 
For the baseline without generated data, the training time is gradually increased to 1,350 epochs (until beginning saturation) in steps of 150 epochs, which corresponds to a maximum of about $5.3\cdot 10^5$ iterations. 
During this evolution, the error on clean data decreases from $10.5$ to $5.5$, and the mCE is reduced to $14.4$.
When using generated data, the size of the total data set is gradually increased from 50,000 (w/o generated images) to 225,000 samples (50,000 real with 175,000 generated images) in steps of 25,000 additional images. 
This stepwise increase leads to a reduction of the error to $3.3$ and to a reduction of the mCE to $10.6$ at 300 training epochs.
In fact, training for more epochs slightly reduces the error and mCE. 
However, by comparing it to the results of the training with generated images, it is evident that significant improvements are induced by utilizing generated data.

\section{Conclusion}
\label{sec:conc}
We propose GenFormer, a generative data augmentation, in order to utilize the inherent robustness of ViTs in the small-scale data domain.
We prove the effectiveness of data augmentation with generated images across various robustness and generalization benchmarks of small datasets including our newly introduced Tiny ImageNetV2, -R and -A test sets.
By combining generative data augmentation with common data augmentation, knowledge distillation and architectural techniques, we show the straightforward applicability and synergistic potential of the proposed method. 
Furthermore, we showcase the consistent improvement under various settings of heavy data limitations by training ViTs on small subsets of Tiny ImageNet.
We conclude that GenFormer encourages a stronger focus on local features in early self-attention layers.
Through the course of this work, we finally close the gap between ViTs and CNNs in terms of accuracy and robustness for limited-data scenarios.

\subsubsection{Acknowledgements} This research was partly funded by Albert and Anneliese Konanz Foundation, the German Research Foundation under grant INST874/9-1 and the Federal Ministry of Education and Research Germany in the project M\textsuperscript{2}Aind-DeepLearning (13FH8I08IA).

\bibliographystyle{splncs04}
\bibliography{main}

% ---- Supplementary ----
\clearpage
\setcounter{page}{1}

\title{GenFormer -- Generated Images are All You Need to Improve Robustness of Transformers on Small Datasets}

\author{Supplementary Material}
\authorrunning{S. Oehri, N. Ebert et al.}
\titlerunning{GenFormer}
\institute{}

\maketitle
\appendix

\setcounter{figure}{5}
\setcounter{table}{4}

\section{Detailed Experimental Settings}
\label{sec:params}

We perform comprehensive classification experiments on the Tiny ImageNet \cite{deng2009imagenet} and CIFAR \cite{krizhevsky2009learning} benchmarks.
Furthermore, we analyze the robustness of classifier networks against various corruptions.
In the following, we explain the different training strategies in more detail.

\textbf{Baseline Setting.} Our training follows Liu et al. \cite{liu2021efficient} and is based on the implementation of Li et al. \cite{li2022locality}.
All models are trained with an input resolution of $224^2$. 
During training, we utilize the AdamW optimizer %\cite{loshchilov2018decoupled} 
with a total batch size of 128, a base learning rate of $5 \times 10^{-4}$ and a weight decay of $0.05$ for 300 epochs.
Furthermore, we employ a cosine decay learning rate schedule % \cite{loshchilov2016sgdr} 
after 20 epochs of linear warm-up.
To avoid overfitting, we adopt RandAugment, %\cite{cubuk2020randaugment}
in addition to standard data augmentation techniques, such as random crop and flip.
For ResNet \cite{he2016deep} we apply some exceptions, following the original training strategy proposed by He et al.
For the optimizer, we employ SGD with a weight decay of $5 \times 10^{-4}$ and a base learning rate of $0.1$.
All classifier trainings are performed on a single Nvidia A100 GPU.

\textbf{Additional Methods.} In the experiments listed in Table \ref{tab:sota}, we use additional strategies during training. In particular, for the combination of CutMix \cite{yun2019cutmix} and Mixup \cite{zhang2018mixup} we stick to the settings of Touvron et al. \cite{touvron2021training}. We apply both on top of the baseline settings with a switching probability of 50\% and the hyperparameters $\alpha_{CutMix}=1.0$ and $\alpha_{Mixup}=0.8$. 
Similarly, AugMix \cite{hendrycks2019augmix} is applied on top of our baseline with a severity of 10. 
In the experiments using Locality Guidance \cite{li2022locality}, we first train ResNet-56 as the CNN teacher for 300 epochs solely on the real dataset at the native Tiny ImageNet input resolution of $64^2$.
Subsequently, the training of the transformer models is conducted utilizing the teacher network for Locality Guidance.
We keep the training settings as proposed by Li et al. \cite{li2022locality}, except for the reduction of the guidance loss weighting $\beta$ from $2.5$ to $0.5$ only when training with generated data in order to enhance the impact of the generative data augmentation.
For test time augmentation (TTA), we follow the default parameters of MEMO  \cite{zhang2022memo}, except that we set the learning rate to $10^{-5}$ and select AdamW as optimizer for the update steps. MEMO is then applied to the already trained model.

\begin{figure*}[h]
    \centering
    \begin{subfigure}[b]{0.44\textwidth}
         \centering
         \includegraphics[width=\textwidth]{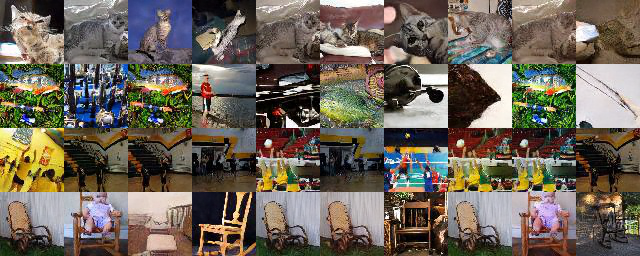}
         \caption{Images generated by the EDM trained on 5\% of Tiny ImageNet}
    \end{subfigure}
     \begin{subfigure}[b]{0.44\textwidth}
         \centering
         \includegraphics[width=\textwidth]{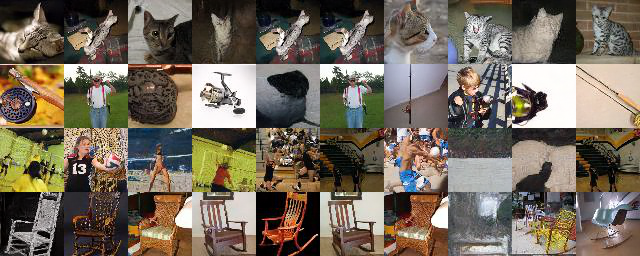}
         \caption{Images generated by the EDM trained on 10\% of Tiny ImageNet}
    \end{subfigure}
     \begin{subfigure}[b]{0.44\textwidth}
         \centering
         \includegraphics[width=\textwidth]{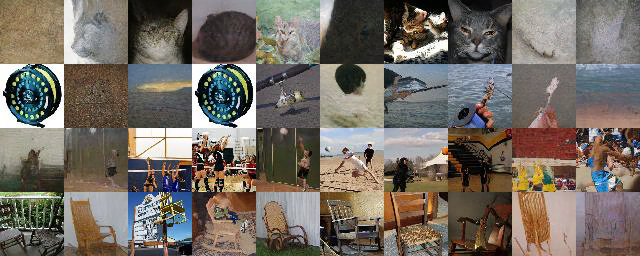}
         \caption{Images generated by the EDM trained on 20\% of Tiny ImageNet}
    \end{subfigure}
     \begin{subfigure}[b]{0.44\textwidth}
         \centering
         \includegraphics[width=\textwidth]{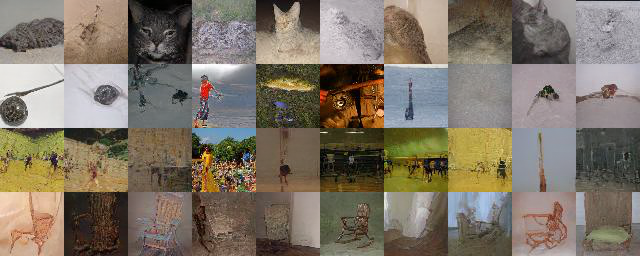}
         \caption{Images generated by the EDM trained on 50\% of Tiny ImageNet}
    \end{subfigure}
     \begin{subfigure}[b]{0.88\textwidth}
         \centering
         \includegraphics[width=\textwidth]{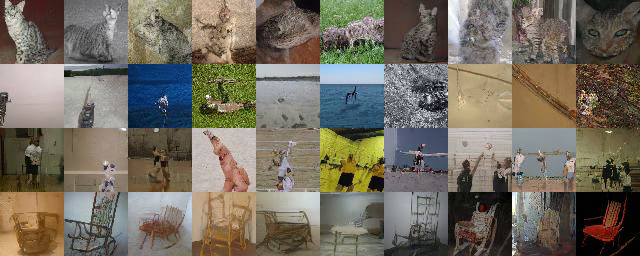}
         \caption{Images generated by the EDM trained on 100\% of Tiny ImageNet}
    \end{subfigure}
    \caption{Uncurated generated images of the EDM \cite{Karras2022edm} diffusion model trained with \{5, 10, 20, 50, 100\}\% of Tiny ImageNet \cite{le2015tiny}. Depicted are the classes \textit{Egyptian cat}, \textit{reel}, \textit{volleyball} and \textit{rocking chair} (line-by-line, from top to bottom).}
    \label{fig:gen_tinyin}
\end{figure*}

\begin{figure*}[h]
\centering
  \includegraphics[width=.88\linewidth]{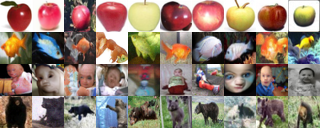}  
  \caption{Uncurated generated images of the EDM \cite{Karras2022edm} diffusion model trained on CIFAR-100 \cite{krizhevsky2009learning}. Depicted are the classes \textit{apple}, \textit{aquarium\_fish}, \textit{baby} and \textit{bear} (line-by-line, from top to bottom).}
  \label{fig:gen_c100}
\end{figure*}

\begin{figure*}[h]
    \centering
    \begin{subfigure}{0.88\textwidth}
         \centering
         \includegraphics[width=\textwidth]{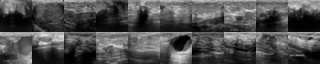}
         \caption{Generated images of BreastMNIST. Depicted are the classes \textit{malignant} and \textit{normal/benign} (line-by-line, from top to bottom).}
         \label{fig:breast_samples}
    \end{subfigure}
    \hfill
    \\
    \begin{subfigure}{0.88\textwidth}
         \centering
         \includegraphics[width=\textwidth]{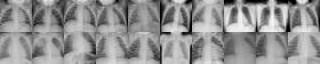}
         \caption{Generated images of PneumoniaMNIST. Depicted are the classes \textit{normal} and \textit{pneumonia} (line-by-line, from top to bottom).}
         \label{fig:pneumonia_samples}
    \end{subfigure}
    \hfill
    \\
    \begin{subfigure}{0.88\textwidth}
         \centering
         \includegraphics[width=\textwidth]{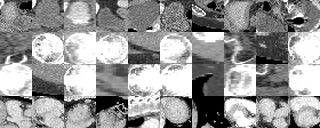}
         \caption{Generated images of OrganSMNIST. Depicted are the classes \textit{bladder}, \textit{femur-left}, \textit{femur-right} and \textit{heart} (line-by-line, from top to bottom).}
         \label{fig:organ_samples}
    \end{subfigure}
    \hfill
        \caption{Uncurated generated images of the EDM \cite{Karras2022edm} diffusion model trained on different datasets of the MedMNIST \cite{medmnistv2} collection.}
        \label{fig:gen_med}
\end{figure*}

\begin{figure*}[ht]
\centering
  \includegraphics[width=.88\linewidth]{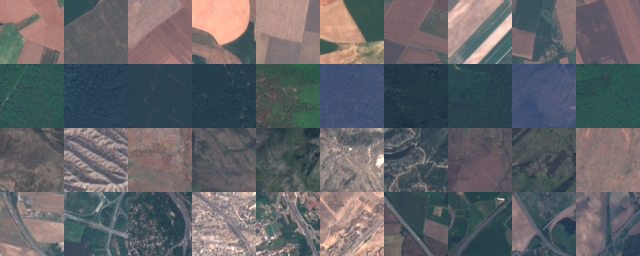}  
  \caption{Uncurated generated images of the EDM \cite{Karras2022edm} diffusion model trained on EuroSAT \cite{helber2019eurosat}. Depicted are the classes \textit{Annual Crop}, \textit{Forest}, \textit{Herbaceous Vegetation} and \textit{Highway} (line-by-line, from top to bottom).}
  \label{fig:gen_eurosat}
\end{figure*}

\textbf{Generator Training.} For training the EDM \cite{Karras2022edm} diffusion model on Tiny ImageNet subsets we used the ImageNet $64^2$ default configuration while adjusting the training duration to \{13000, 25000, 50000, 125000, 250000\} k-imgs for the subsets \{5, 10, 20, 50, 100\}\% respectively, in order to accommodate for the different training set sizes. Additionally, the GPU batch-size was limited to 128. For CIFAR-100, we used the default CIFAR-10 $32^2$ variance preserving (VP) configuration. For the generation of CIFAR-100 and Tiny ImageNet samples, we followed the default generation settings for CIFAR-10 and ImageNet, with the exception of sampling steps, which were set to 54 for CIFAR-100, and 768 for Tiny ImageNet. The training of EDM diffusion models and sample generation are carried out on a node of eight Nvidia A100 GPUs. Uncurated generated samples of Tiny ImageNet, CIFAR-100, MedMNIST, and EuroSAT are depicted in the Figures \ref{fig:gen_tinyin}, \ref{fig:gen_c100}, \ref{fig:gen_med}, and \ref{fig:gen_eurosat} respectively.

\section{Further Studies of Limited-Data on Tiny ImageNet}
\label{sec:appendix_limited_data}

\begin{figure*}[!t]
\centering
  \includegraphics[width=.99\linewidth]{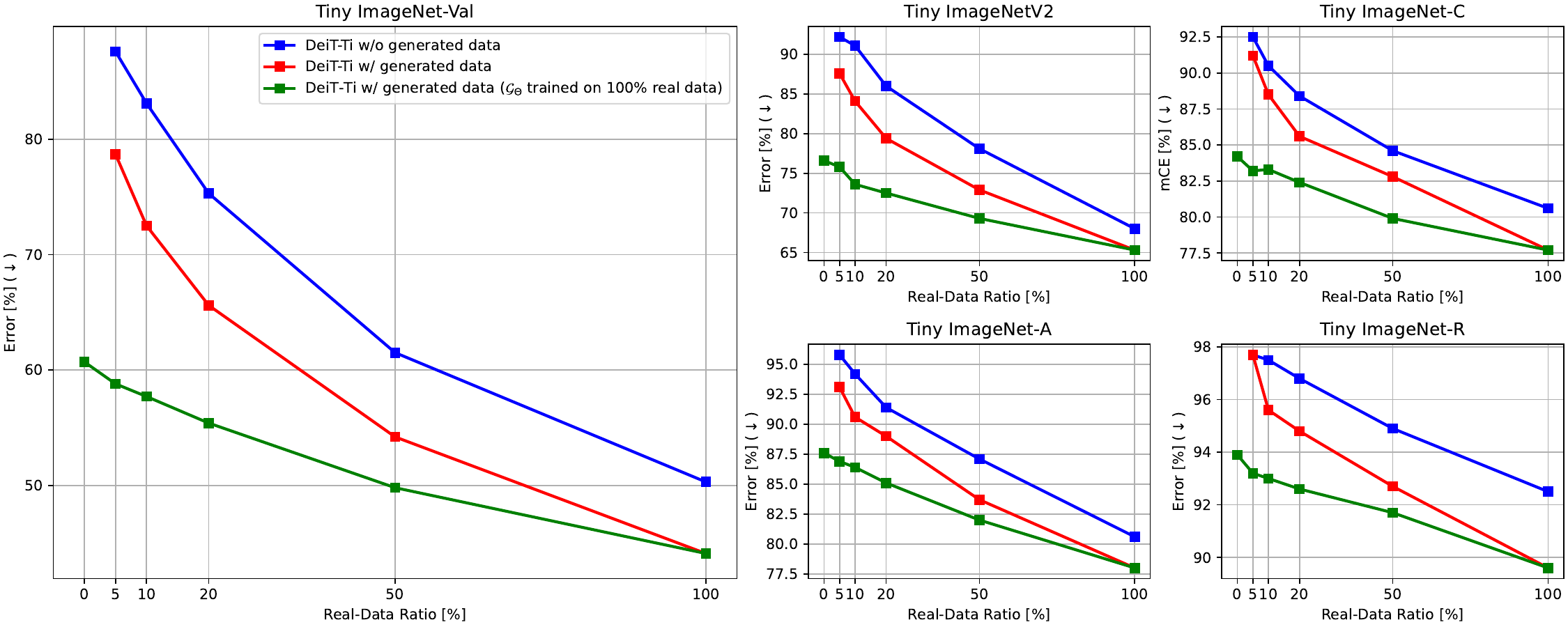}  
  \caption{Analysis of the influence of different amounts of training data on the accuracy as well as the robustness of DeiT-Ti \cite{touvron2021training}. The networks are trained with \{5, 10, 20, 50, 100\}\% of Tiny ImageNet \cite{le2015tiny}. We add 100,000 generated images of a diffusion model \cite{Karras2022edm} trained on the same set (\textcolor{red}{red line}) and on all data (\textcolor{ForestGreen}{green line}) to the respective set of training data. A Real-Data Ratio of 0\% corresponds to a training using only generated images.}
  \label{fig:subset+}
  \vspace{+4mm}
\end{figure*}

\begin{figure*}[h]
    \centering
    \begin{subfigure}{0.49\textwidth}
         \centering
         \includegraphics[width=\textwidth]{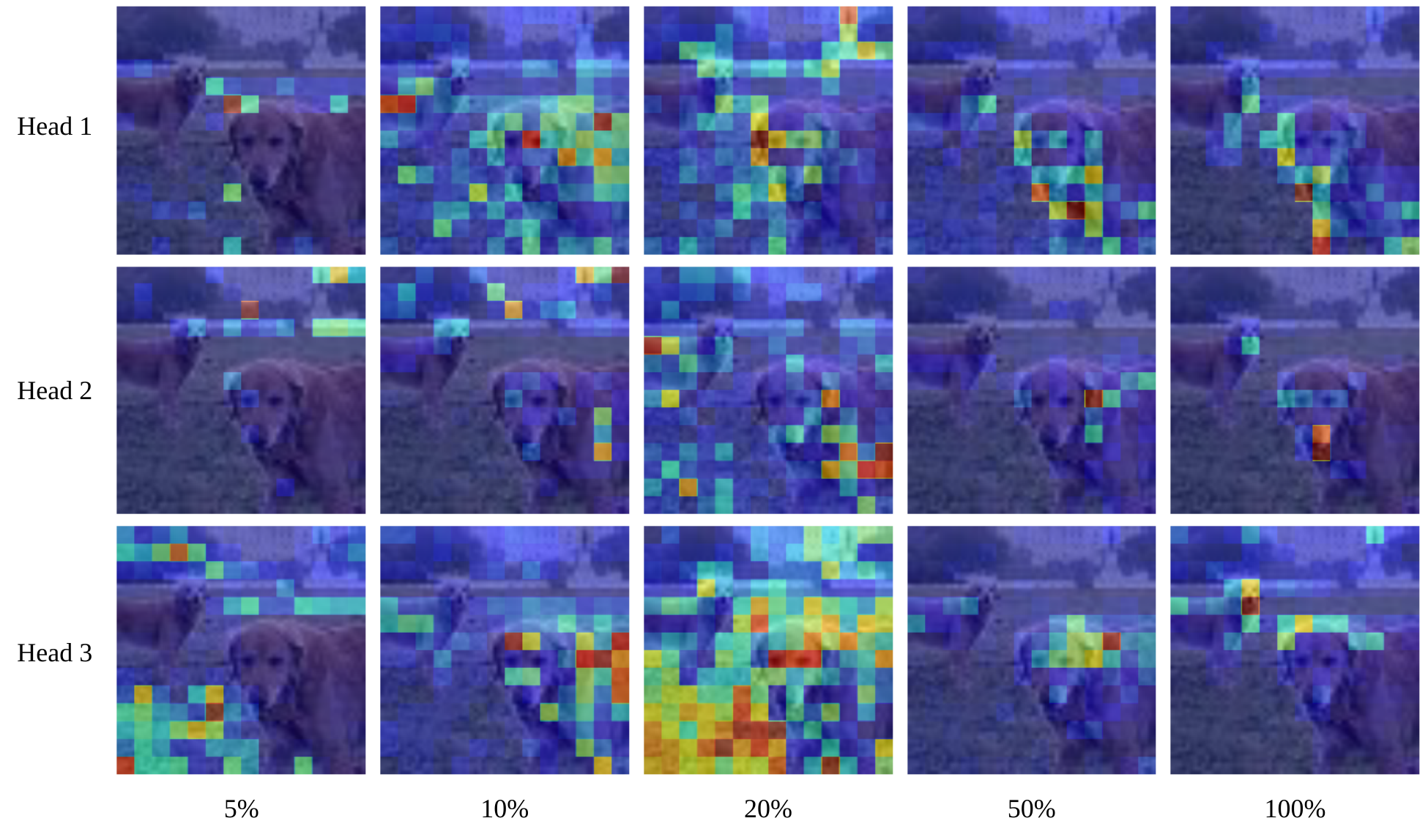}
         \caption{Attention on Tiny ImageNet without generated data}
         \label{fig:attn_tin}
    \end{subfigure}
    \hfill
    \begin{subfigure}[b]{0.49\textwidth}
         \centering
         \includegraphics[width=\textwidth]{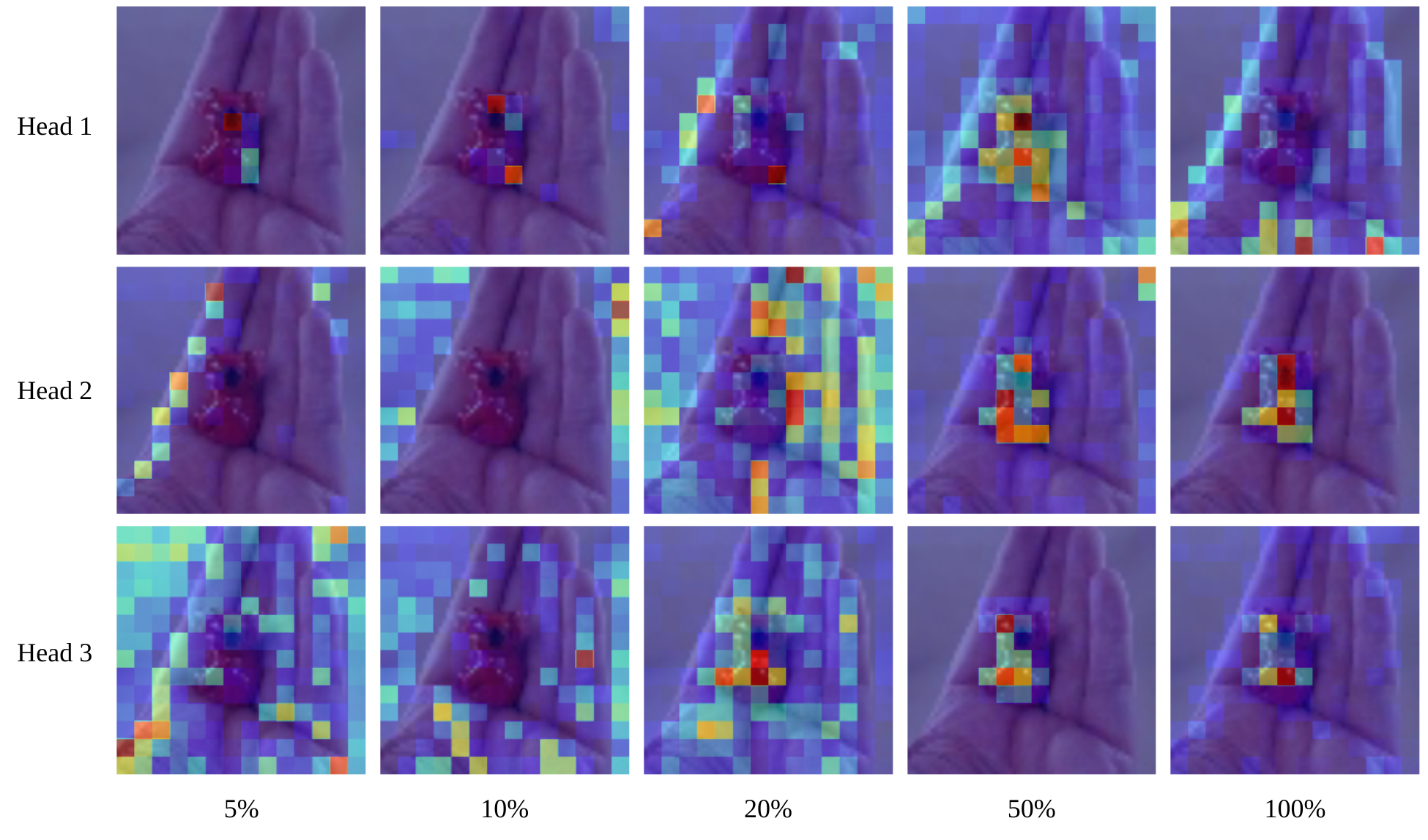}
         \caption{Attention on Tiny ImageNet-R without generated data}
         \label{fig:attn_tinr}
    \end{subfigure}
    \\
    \begin{subfigure}[b]{0.49\textwidth}
         \centering
         \includegraphics[width=\textwidth]{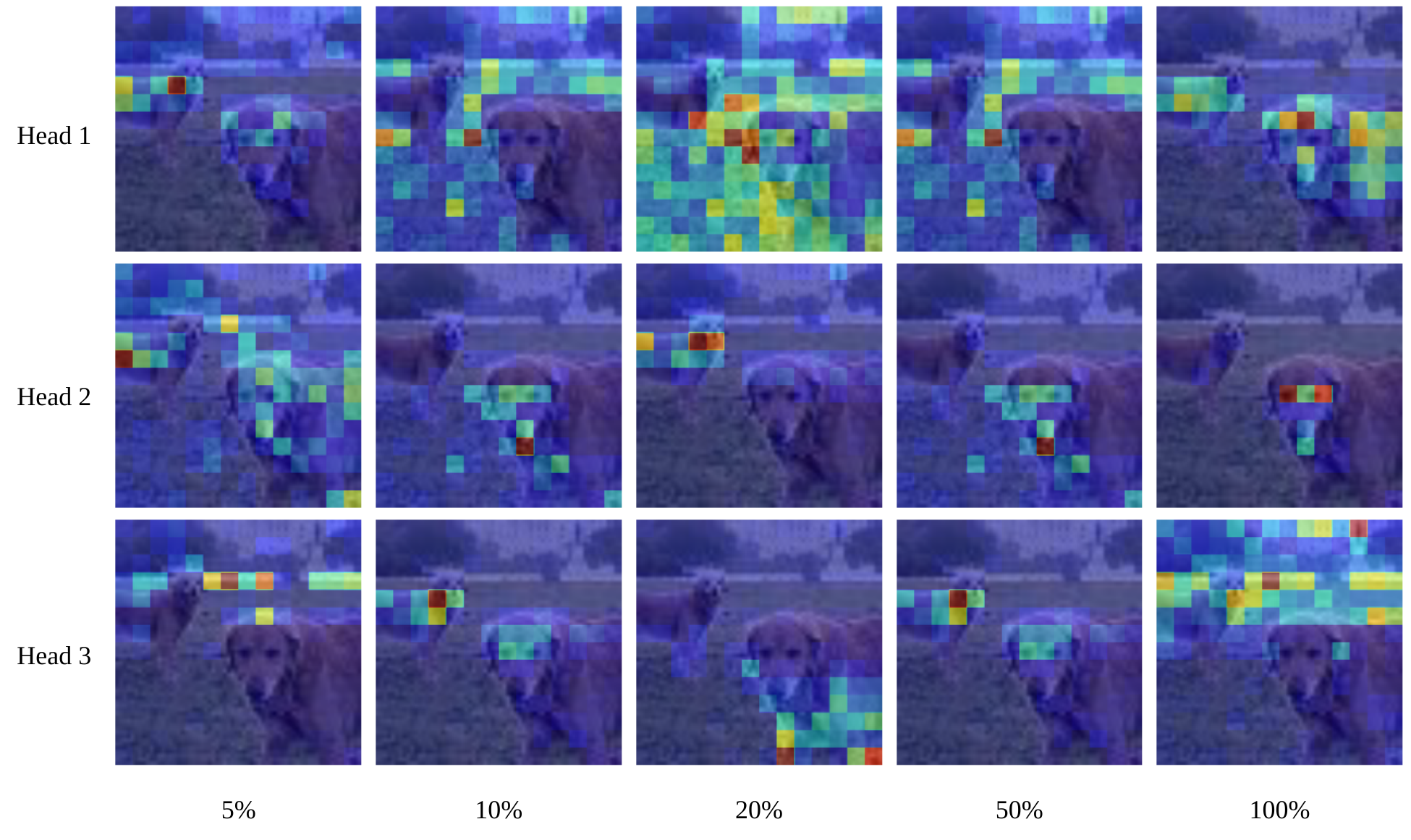}
         \caption{Attention on Tiny ImageNet with generated data}
         \label{fig:attn_tin_syn}
    \end{subfigure}
    \hfill
    \begin{subfigure}[b]{0.49\textwidth}
         \centering
         \includegraphics[width=\textwidth]{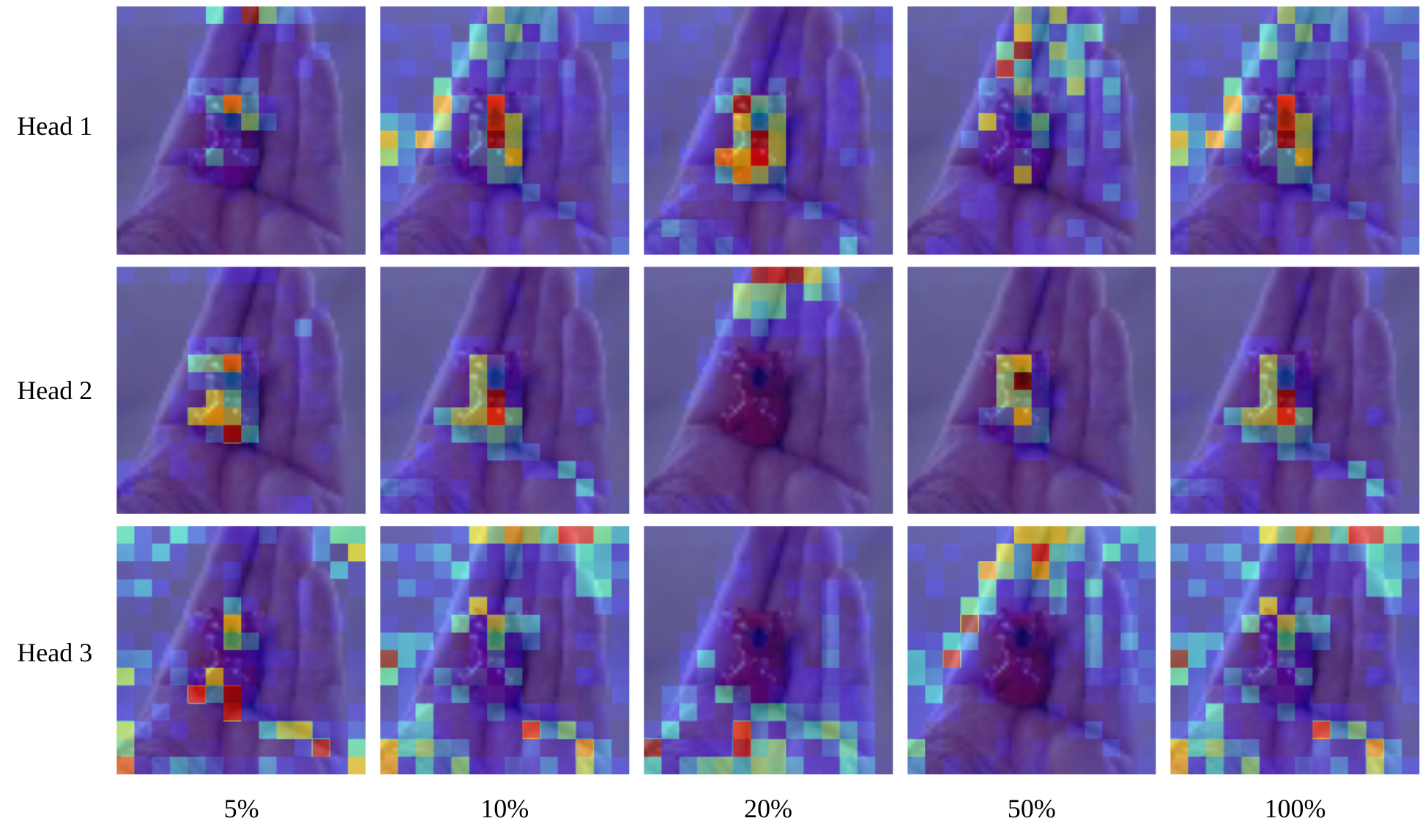}
         \caption{Attention on Tiny ImageNet-R with generated data}
         \label{fig:attn_tinr_syn}
    \end{subfigure}
        \caption{Analysis of the influence of different amounts of training data on the attention-maps. The networks are trained with \{5, 10, 20, 50, 100\}\% of Tiny ImageNet \cite{le2015tiny}. We also add 100k generated images to each subset.}
        \label{fig:three graphs}
\end{figure*}

Extending the experiments described in Section \ref{sec:limited}, we investigate the potential improvements that can be achieved for individual Tiny ImageNet subsets (5\%, 10\%, 20\% and 50\%) by extending each subset with high-variance generated data. To enable this, we generate a dataset of 100,000 images with an EDM \cite{Karras2022edm} diffusion model trained with all images of Tiny ImageNet. We then train DeiT-Ti on the above subsets in combination with the generated data. Additionally, we perform a training using only the generated data (corresponding to a Real-Data Ratio of 0\%). The resulting performance on the validation set and the individual benchmarks is shown graphically in Figure \ref{fig:subset+} (green line). Furthermore we analyze the performance of DeiT-Ti without additional data (blue line), as well as scenarios where additional data was generated using a diffusion model trained exclusively with the respective subsets (red line).

The results illustrate the significant improvement achieved by incorporating images of a generative network with access to a higher amount of real images.
When trained exclusively with generated images ($60.7$), the clean error increases by only $10.4$ compared to the model trained only with the real dataset ($50.3$) and even drops below the error of the network trained with 50\% of the real images ($61.5$).
These benefits notably extend to the robustness benchmarks, with the greater data diversity providing a clear advantage.
For example, it is possible to achieve a mCE of less than $85.0$ for Tiny ImageNet-C without using real data.

\section{Futher Studies on CIFAR-100}
\label{sec:appendix_big_models}

\begin{table}[!t]
\scriptsize
\caption{Scalability of GenFormer with different model sizes of DeiT \cite{touvron2021training} and RVT \cite{mao2022towards} on CIFAR-100 \cite{krizhevsky2009learning} and the corresponding robustness benchmark \cite{hendrycks2018benchmarking}. }
\label{tab:sizes}
\centering
\begin{tabular}{ll|c|ll}
\hline
\multicolumn{2}{c|}{\multirow{2}{*}{\textbf{Method}}} & \multirow{2}{*}{\textbf{Params}} & \multicolumn{1}{c}{\textbf{C-100}}  & \multicolumn{1}{c}{\textbf{C-100-C}} \\
\multicolumn{2}{c|}{}  & & \multicolumn{1}{c}{\textbf{err.}} & \multicolumn{1}{c}{\textbf{mCE}}   \\ \hline
\multirow{2}{*}{DeiT-Ti \cite{touvron2021training}}          & w/o ours          & \multirow{2}{*}{5.5M}           & 35.3          & 51.8           \\
 & w/ ours           &    & 24.7 \color{ForestGreen}{(-10.6)}          & 39.4 \color{ForestGreen}{(-12.4)}          \\ \cline{3-5}
\multirow{2}{*}{RVT-Ti \cite{mao2022towards}}           & w/o ours          & \multirow{2}{*}{8.3M}           & 18.1          & 31.1           \\
& w/ ours           & & 15.3 \color{ForestGreen}{(-2.8)}         & 27.3 \color{ForestGreen}{(-3.8)}          \\ \hline
\multirow{2}{*}{DeiT-S \cite{touvron2021training}}           & w/o ours          & \multirow{2}{*}{21.6M}          & 35.6          & 52.2           \\
& w/ ours           &   & 22.2 \color{ForestGreen}{(-13.4)}         & 37.1 \color{ForestGreen}{(-15.1)}          \\ \cline{3-5}
\multirow{2}{*}{RVT-S \cite{mao2022towards}}            & w/o ours          & \multirow{2}{*}{21.8M}          & 17.6          & 30.3           \\
& w/ ours           &   & 15.4 \color{ForestGreen}{(-2.2)}         & 27.0 \color{ForestGreen}{(-3.3)}          \\ \hline
\multirow{2}{*}{DeiT-B \cite{touvron2021training}}           & w/o ours          & \multirow{2}{*}{85.7M}          & 39.6          & 55.9           \\
& w/ ours           &  & 21.2 \color{ForestGreen}{(-18.4)}         & 35.2 \color{ForestGreen}{(-20.7)}          \\ \cline{3-5}
\multirow{2}{*}{RVT-B \cite{mao2022towards}}            & w/o ours          & \multirow{2}{*}{85.5M}          & 34.7          & 52.1           \\
& w/ ours           &  & 20.1 \color{ForestGreen}{(-14.6)}         & 36.3 \color{ForestGreen}{(-15.8)}          \\ \hline
\end{tabular}
\end{table}

In addition to our analysis from Section \ref{sec:cifar}, we conduct further experiments on the CIFAR-100 \cite{krizhevsky2009learning} dataset.
On one hand, we investigate the scalability of our proposed GenFormer approach, by training larger models of around 20M and 90M parameters.
On the other hand, we combine GenFormer with multiple established methods leading to significant improvements and achieving state-of-the-art performance.
When applying our GenFormer approach we follow the same strategy as in Section \ref{sec:cifar} and extend the real dataset with 100k additional images generated by an EDM \cite{Karras2022edm} diffusion model, which is trained on CIFAR-100.

\textbf{Model Scalability.} For the scalability experiment, we select the small and base variants of DeiT \cite{touvron2021training} and RVT \cite{mao2022towards} and train on CIFAR-100 with the baseline settings.
The results (see Table \ref{tab:sizes}) highlight the advantage of applying our GenFormer approach when training larger models from scratch.
In the case of DeiT, we observe signs of overfitting increasing with the amount of model parameters when the training is performed solely on the real dataset.
However, utilizing additional generated data prevents overfitting for all DeiT variants.
This leads to a consistent reduction of up to $-18.4$ and $-20.7$ for the clean and mean corruption error.
The absolute best performance is achieved by DeiT-B with an improvement of $-3.5$ and $-4.2$ compared to DeiT-Ti, when trained with our GenFormer approach.
Similar to DeiT, our proposed generative data augmentation also results in consistent performance gains for all RVT variants.
However, in contrast to the DeiT architecture, we observe a strong overfitting behavior only for the base variant of RVT, which is greatly reduced with the help of our method.
We conclude that the proposed method scales with larger models and combats the need for additional data to reduce overfitting, resulting in performance gains.

\begin{table}[ht]
\scriptsize
\caption{Evaluation of an optimal training strategy combining GenFormer with RVT \cite{mao2022towards} and established methods CutMix \cite{yun2019cutmix}, Mixup \cite{zhang2018mixup} and MEMO \cite{zhang2022memo}, on CIFAR-100 \cite{krizhevsky2009learning} and the corresponding robustness benchmark \cite{hendrycks2018benchmarking}. }
\label{tab:optimal}
\centering
\begin{tabular}{l|ll}
\hline
\multicolumn{1}{c|}{\multirow{2}{*}{\textbf{Model}}}                     & \multicolumn{1}{c}{\textbf{C-100}}  & \multicolumn{1}{c}{\textbf{C-100-C}} \\
\multicolumn{1}{c|}{}                                                    & \multicolumn{1}{c}{\textbf{err.}} & \multicolumn{1}{c}{\textbf{mCE}}   \\ \hline
RVT-Ti \cite{mao2022towards}                                             & 18.1          & 31.1           \\
+ GenFormer (ours)                                                       & 15.3 \color{ForestGreen}{(-2.8)}         & 27.3 \color{ForestGreen}{(-3.8)}          \\
\hspace{2mm} + CutMix \cite{yun2019cutmix} / Mixup \cite{zhang2018mixup} & 13.5 \color{ForestGreen}{(-1.8)}         & 25.2 \color{ForestGreen}{(-2.1)}          \\
\hspace {4.5mm} + MEMO \cite{zhang2022memo}                              & 13.5 ($\pm$0.0)         & 24.6 \color{ForestGreen}{(-0.6)}          \\
\hline
RVT-S \cite{mao2022towards}                                              & 17.6          & 30.3           \\
+ GenFormer (ours)                                                       & 15.4 \color{ForestGreen}{(-2.2)}         & 27.0 \color{ForestGreen}{(-3.3)}          \\
\hspace{2mm} + CutMix \cite{yun2019cutmix} / Mixup \cite{zhang2018mixup} & 13.2 \color{ForestGreen}{(-2.2)}         & 24.5 \color{ForestGreen}{(-2.5)}          \\
\hspace {4.5mm} + MEMO \cite{zhang2022memo}                              & 13.4 \color{BrickRed}{(+0.2)}            & 24.0 \color{ForestGreen}{(-0.5)}          \\
\hline
\end{tabular}
\end{table}

\textbf{Optimized Training.} 
We further investigate an optimal strategy by combining the proposed GenFormer approach with established methods.
For this purpose, we select RVT-Ti and -S \cite{mao2022towards} as the base models, chosen due to their robust architecture.
We then apply our generative data augmentation and combine it with the conventional data augmentation techniques CutMix \cite{yun2019cutmix} and Mixup \cite{zhang2018mixup} as well as MEMO \cite{zhang2022memo}.
We observe a steady improvement by combining all methods resulting in state-of-the-art clean and mean corruption errors for small models trained from scratch.
The strong performance of the RVT-Ti model is particularly noteworthy, performing almost as well as the RVT-S model, however, at a third of its size.

\section{Attention Analysis}
\label{sec:attention_maps}

To understand why our method especially performs well for the transformer architecture, we conduct further experiments on the Tiny ImageNet dataset focusing on the self-attention mechanism. When looking at the attention maps, it is noticeable (see Figure \ref{fig:attn_tin}) that the attention maps of the last layer in DeiT-Ti \cite{touvron2021training} are not focused on the objects in the settings of limited amount of training data. It becomes evident that without the GenFormer approach, at least 50,000 real images are required to help the network directing its attention to recognizable objects. Conversely, when processing only 5\% of the real data augmented with 100,000 generated images, the network shows the ability to highlight objects belonging to the respective class (see Figure \ref{fig:attn_tin_syn}). This pattern remains consistent in the case of the Tiny ImageNet-R example. Here, a recognizable focus on the critical object only becomes apparent when 50\% of the real data is used as shown in Figure \ref{fig:attn_tinr}. Remarkably, the introduction of additional generated data improves the ability of the network to focus on the target object even in scenarios with limited real data (Figure \ref{fig:attn_tinr_syn}).

\begin{figure*}[h]
    \centering
    \begin{subfigure}{0.49\textwidth}
         \centering
         \includegraphics[width=\textwidth]{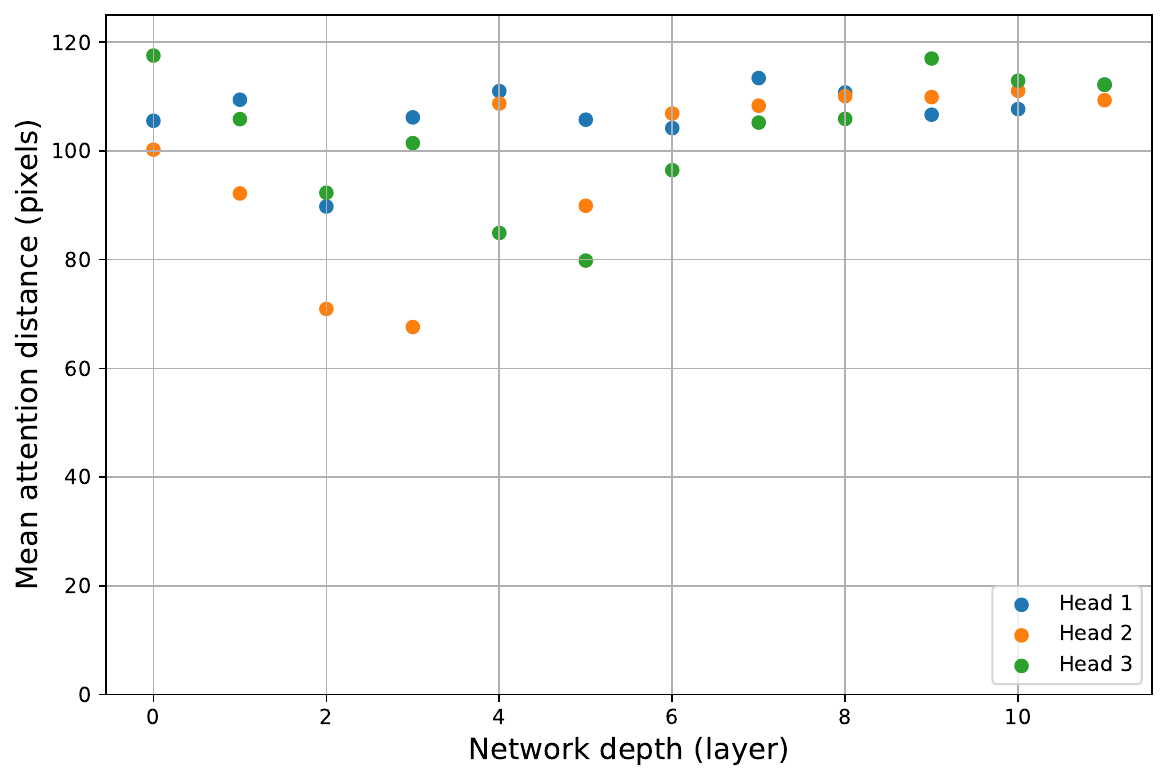}
         \caption{Attention Distance with 100\% of Tiny ImageNet w/o generated data}
         \label{fig:attn_dist_100}
    \end{subfigure}
    \hfill
    \begin{subfigure}[b]{0.49\textwidth}
         \centering
         \includegraphics[width=\textwidth]{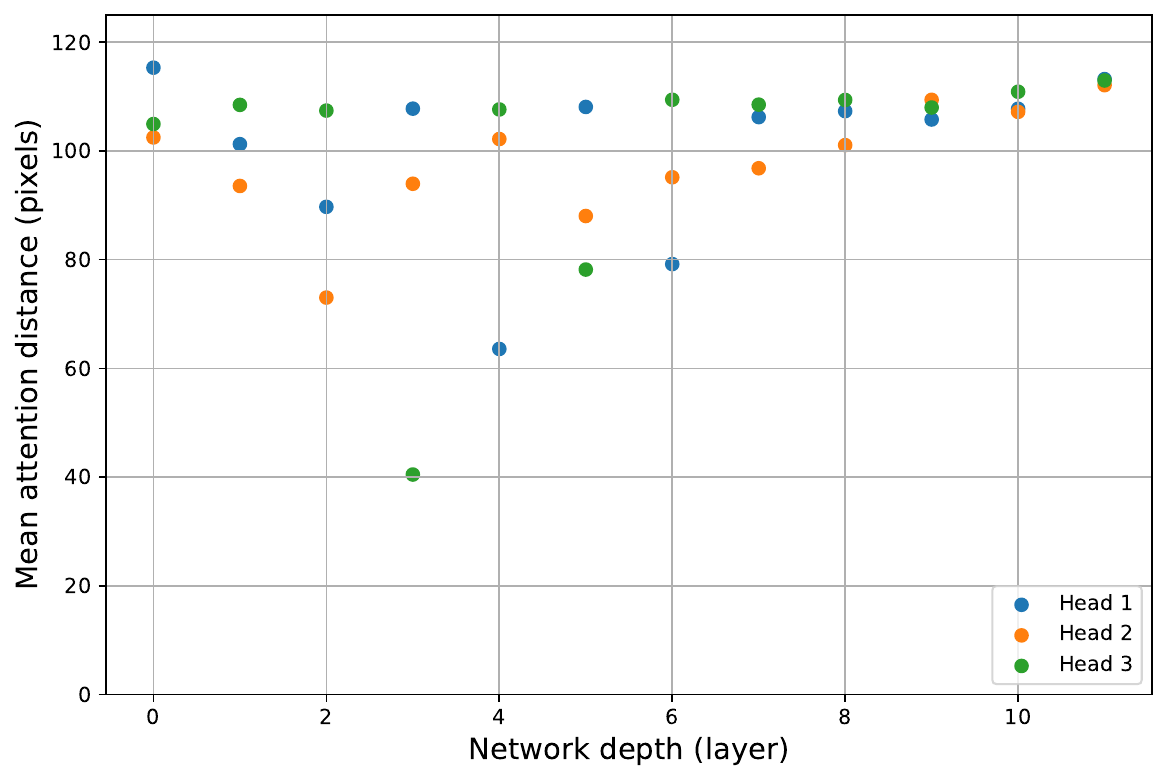}
         \caption{Attention Distance with 100\% of Tiny ImageNet w/ generated data}
         \label{fig:attn_dist_100_syn}
    \end{subfigure}
    \\
    \begin{subfigure}[b]{0.49\textwidth}
         \centering
         \includegraphics[width=\textwidth]{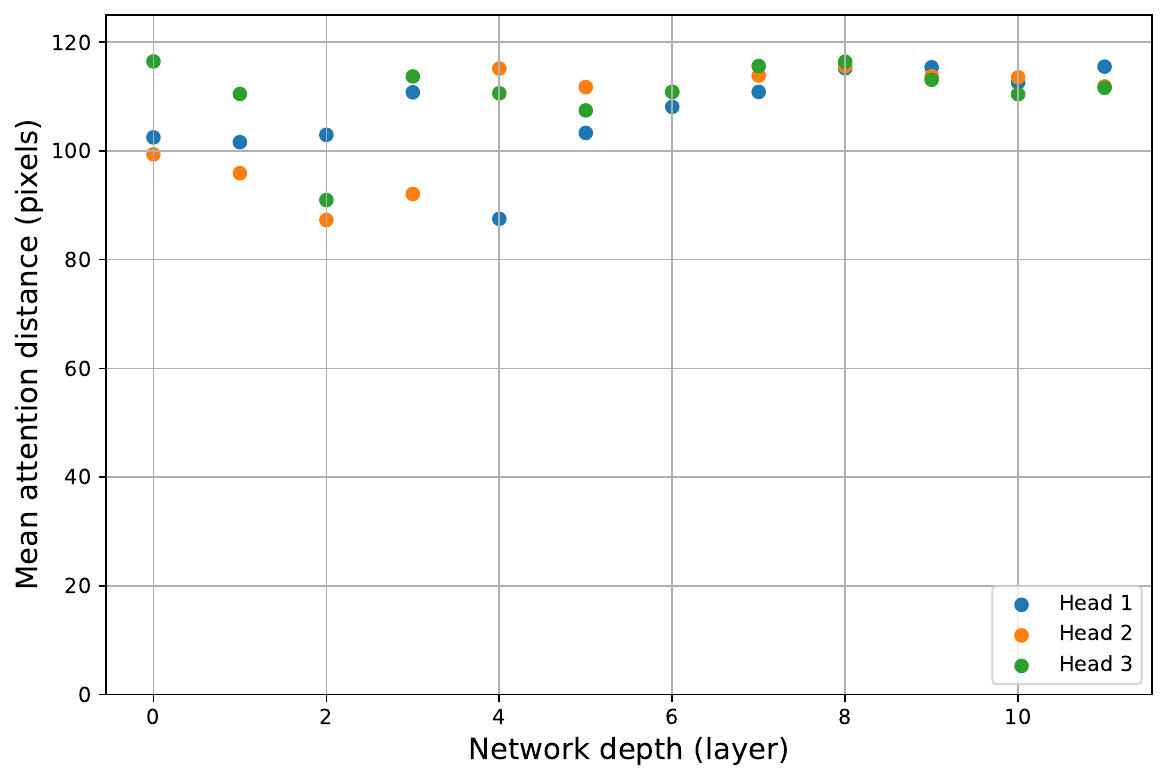}
         \caption{Attention Distance with 50\% of Tiny ImageNet w/o generated data}
         \label{fig:attn_dist_50}
    \end{subfigure}
    \hfill
    \begin{subfigure}[b]{0.49\textwidth}
         \centering
         \includegraphics[width=\textwidth]{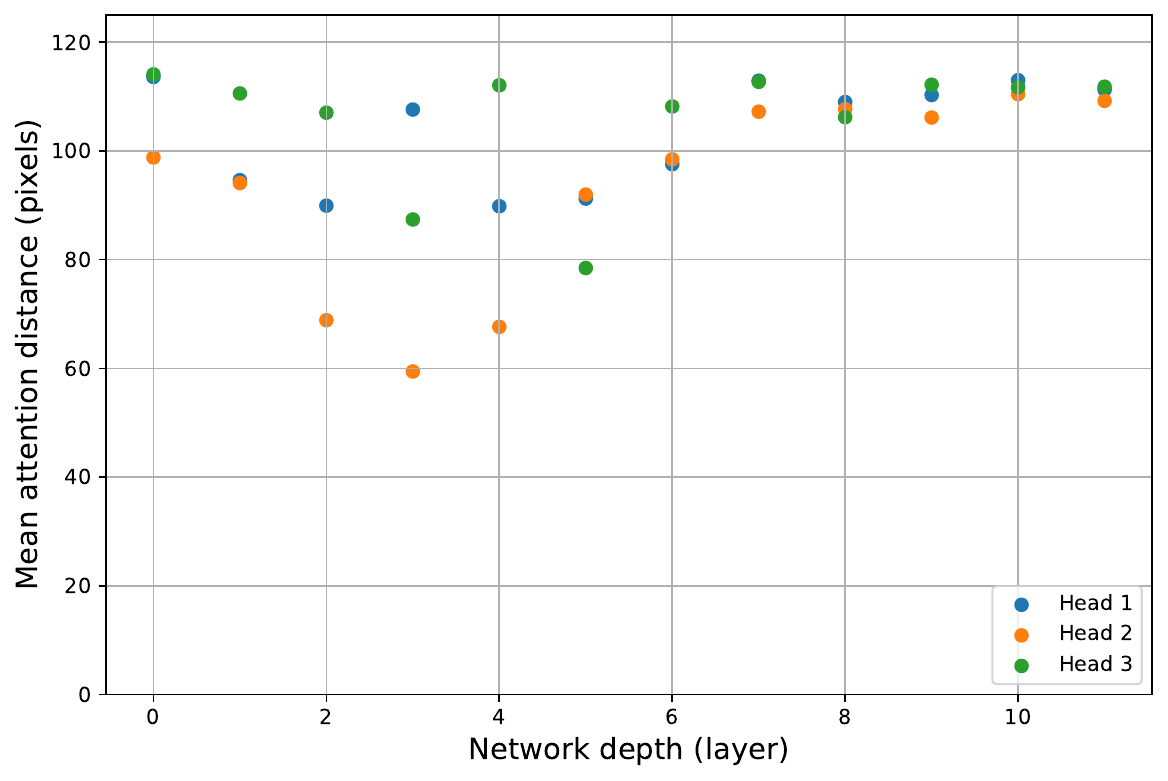}
         \caption{Attention Distance with 50\% of Tiny ImageNet w/ generated data}
         \label{fig:attn_dist_50_syn}
    \end{subfigure}
    \\
    \begin{subfigure}[b]{0.49\textwidth}
         \centering
         \includegraphics[width=\textwidth]{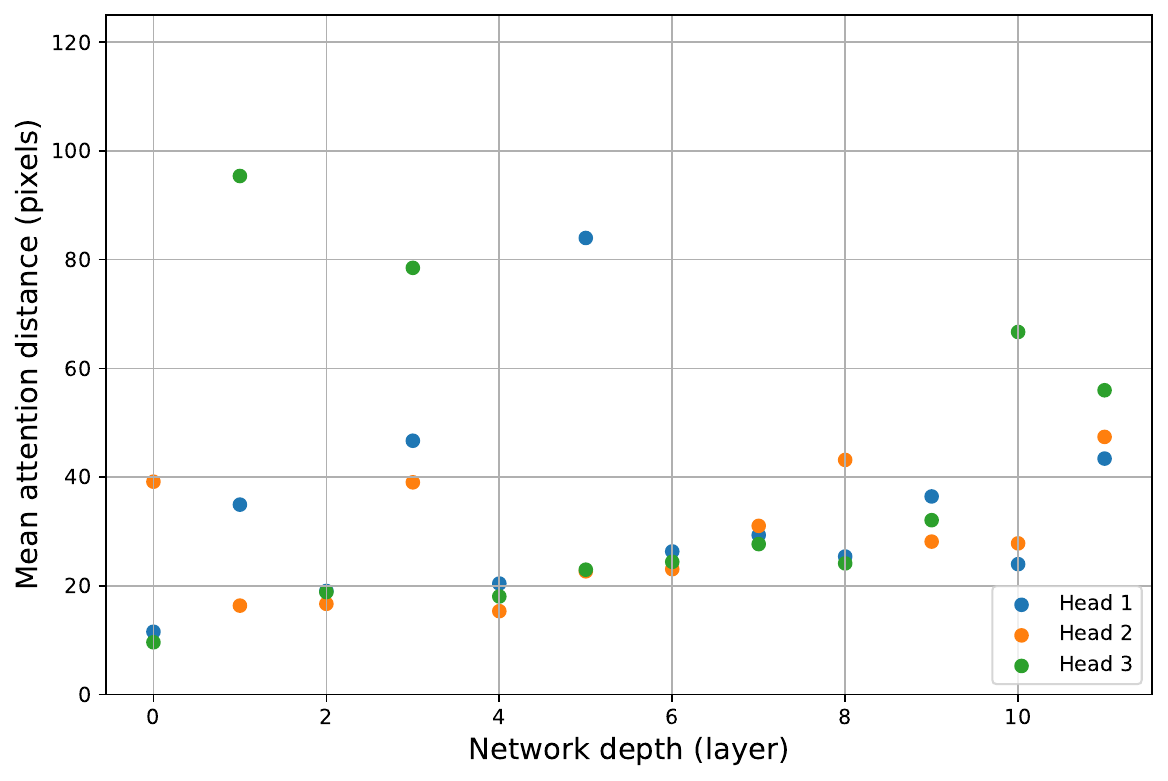}
         \caption{Attention Distance with 100\% of Tiny ImageNet and Locality Guidance w/o generated data}
         \label{fig:attn_dist_local}
    \end{subfigure}
    \hfill
    \begin{subfigure}[b]{0.49\textwidth}
         \centering
         \includegraphics[width=\textwidth]{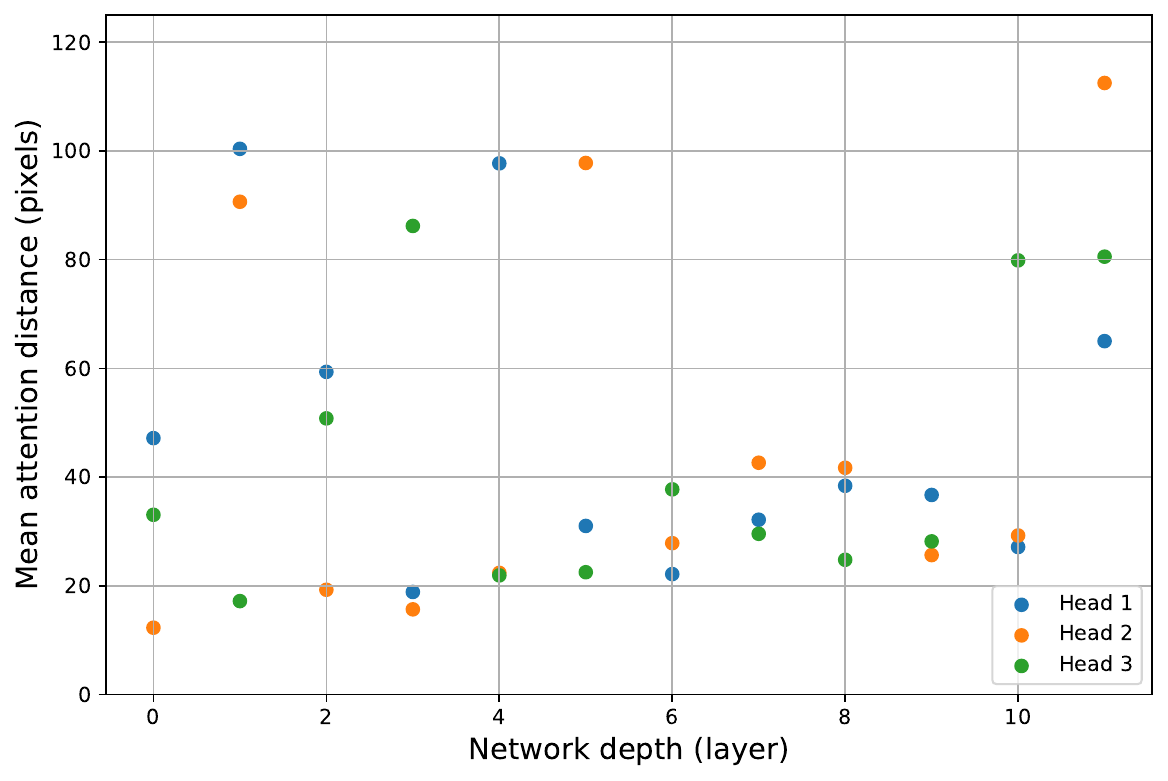}
         \caption{Attention Distance with 100\% of Tiny ImageNet and Locality Guidance w/ generated data}
         \label{fig:attn_dist_local_syn}
    \end{subfigure}
        \caption{Analysis of the mean attention distances of a DeiT-Ti \cite{touvron2021training} trained on Tiny ImageNet \cite{le2015tiny} with 100\% and 50\% of the data as well as with the use of Locality Guidance \cite{li2022locality}. Each scenario is conducted with and without the use of 100k additional generated images.}
        \label{fig:attn_dist}
\end{figure*}

In addition to attention maps, we study the mean attention distances of DeiT-Ti for three scenarios: trained (1) with the complete real dataset, (2) with a subset of 50\% of the real dataset and (3) with Locality Guidance \cite{li2022locality} and the complete real dataset. Each scenario is conducted with and without an additional 100k generated images of an EDM\cite{Karras2022edm} diffusion model, which is trained on 100\% for case (1) and (3) and 50\% of Tiny ImageNet for case (2).
When comparing Figure \ref{fig:attn_dist_100} with Figure \ref{fig:attn_dist_100_syn} we observe a reduction of the mean attention distances when applying GenFormer, especially in early layers.
This can be interpreted as the induction of a stronger local bias, which transformer models trained on small-scale datasets are usually lacking.
This effect becomes even stronger the less real data is available, as can be seen in Figure \ref{fig:attn_dist_50} and \ref{fig:attn_dist_50_syn}.
Locality Guidance exhibits a very different behavior.
As proposed by Li et al. \cite{li2022locality} a strong local bias is induced to the transformer model, which becomes evident by very low mean attention distances throughout all layers (see Figure \ref{fig:attn_dist_local}).
However, it rather looks like the characteristics of the CNN teacher are being imposed on the transformer model, whereby the advantageous global nature of the transformer is lost.
By utilizing GenFormer and reducing the impact of the Locality Guidance (see Section \ref{sec:params}) we combine the beneficial characteristics of the CNN teacher and the transformer model.
This results in both, low and high mean attention distances (see Figure \ref{fig:attn_dist_local_syn}), thus leading to a strong locality while preserving the globality.

\section{Test Set Analysis}
\label{sec:appendix_test_sets}
\begin{figure*}[t]
    \centering
    \begin{subfigure}{0.45\textwidth}
    \centering
    \includegraphics[width=\textwidth]{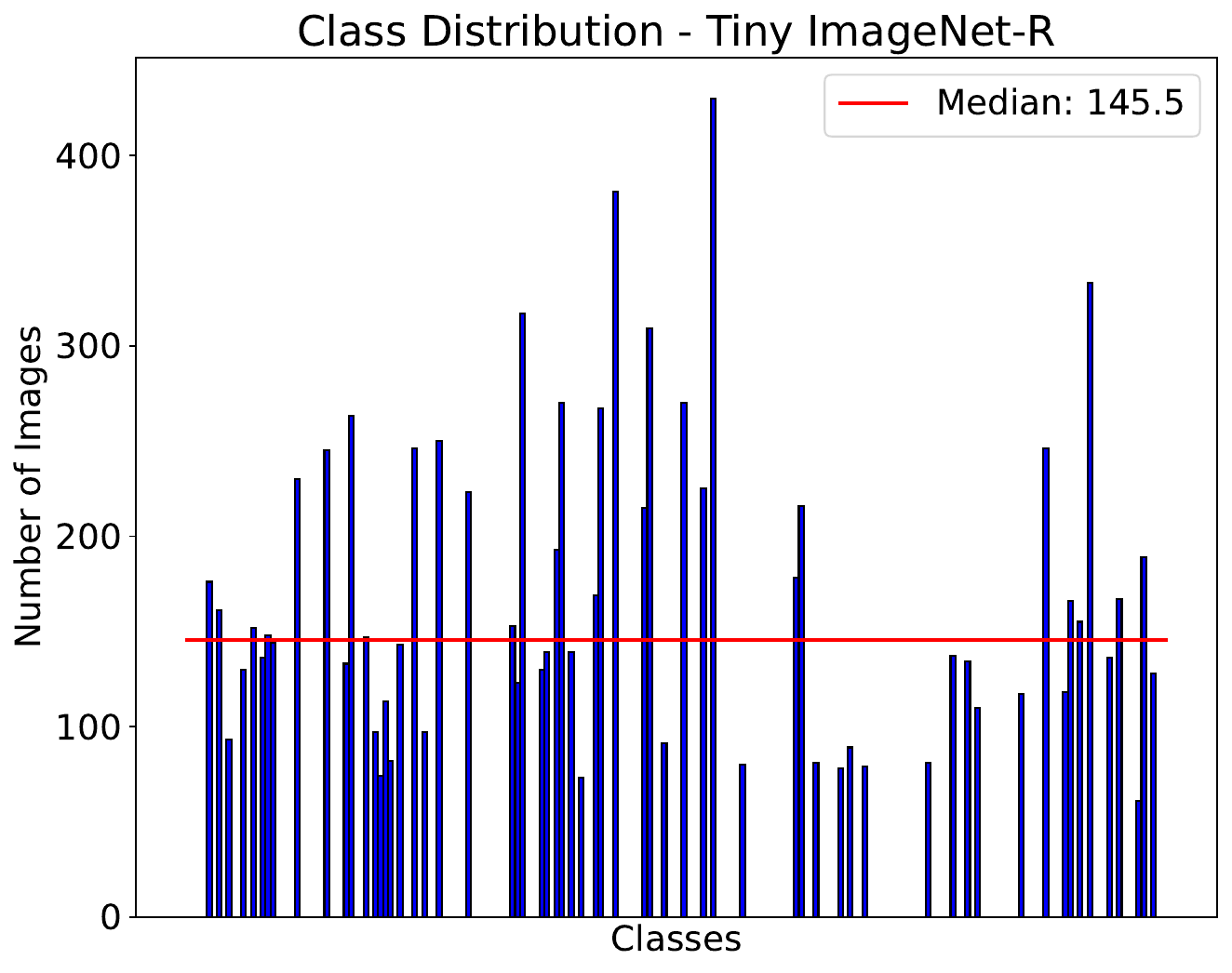}
    \caption{\protect\centering Class distribution of Tiny ImageNet-R test set}
    \label{fig:tin-r_class}
    \end{subfigure}
    \begin{subfigure}{0.45\textwidth}
    \centering
    \includegraphics[width=\textwidth]{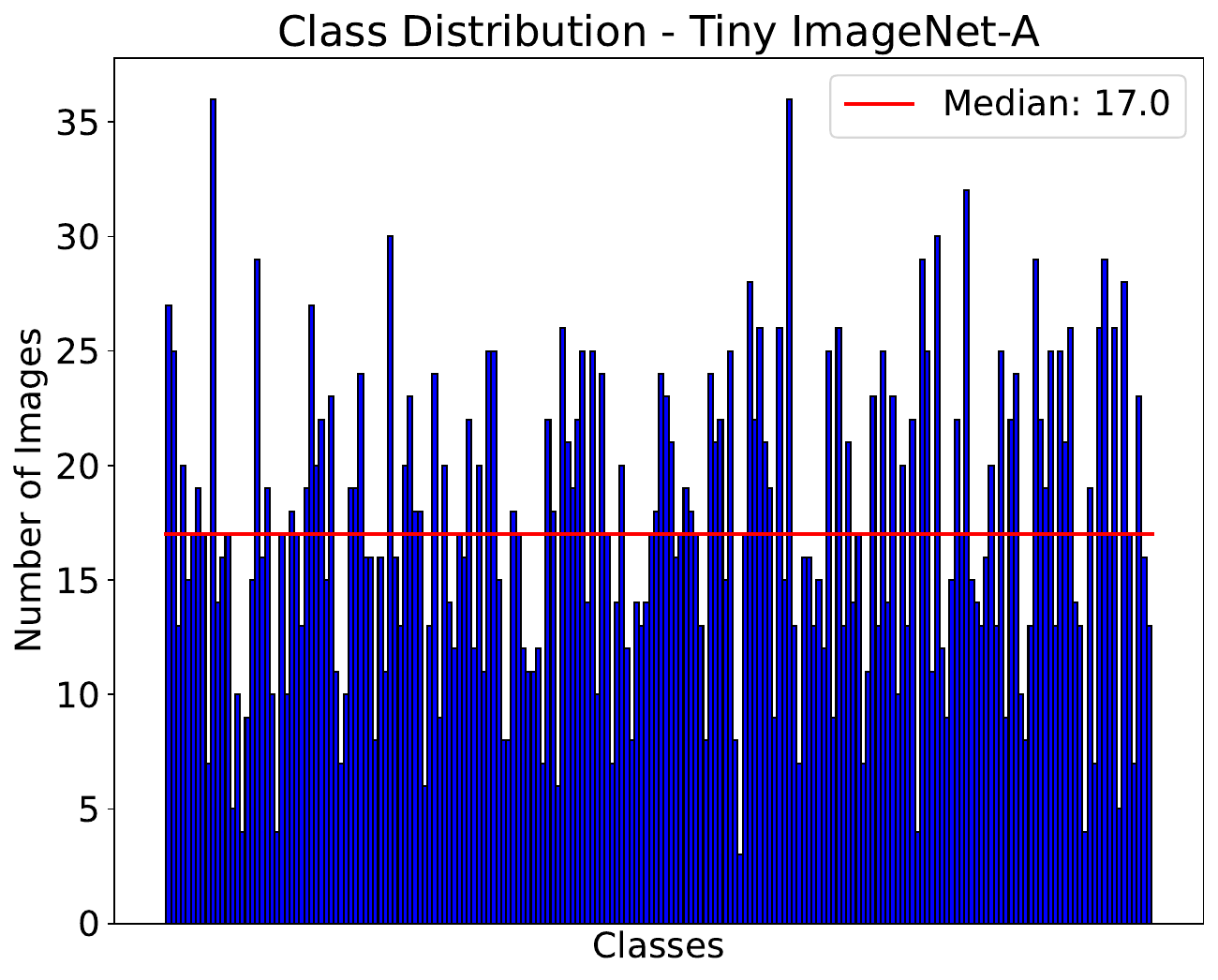}  
    \caption{\protect\centering Class distribution of Tiny ImageNet-A test set.}
    \label{fig:tin-a_class}
    \end{subfigure}
    \caption{Class distributions of our proposed Tiny ImageNet test sets. The \textcolor{red}{red line} corresponds to the median images per class.}
    \label{fig:datasets}
\end{figure*}

\begin{figure*}[h]
    \centering
    \begin{subfigure}{0.99\textwidth}
         \centering
         \includegraphics[width=\textwidth]{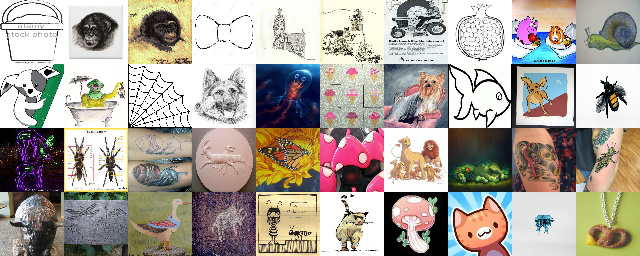}
         \caption{Uncurated random samples of Tiny ImageNet-R test set}
         \label{fig:tin-r_samples}
    \end{subfigure}
    \hfill
    \\
    \begin{subfigure}{0.99\textwidth}
         \centering
         \includegraphics[width=\textwidth]{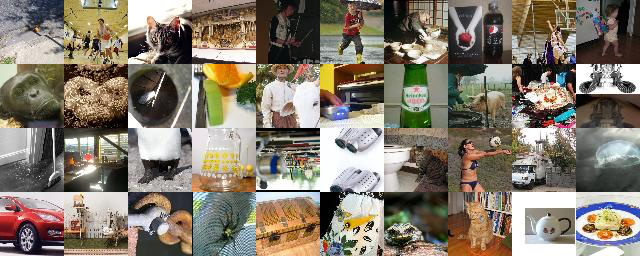}
         \caption{Uncurated random samples of Tiny ImageNet-A test set}
         \label{fig:tin-a_samples}
    \end{subfigure}
    \hfill
    \\
    \begin{subfigure}{0.99\textwidth}
         \centering
         \includegraphics[width=\textwidth]{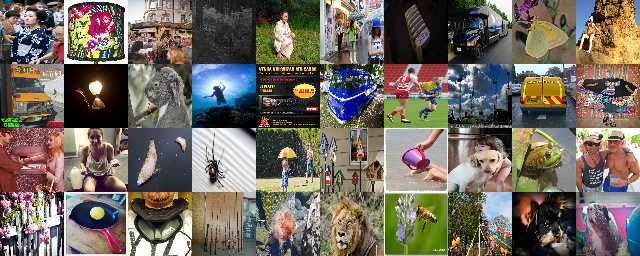}
         \caption{Uncurated random samples of Tiny ImageNetV2 test set}
         \label{fig:tin-v2_samples}
    \end{subfigure}
    \hfill
        \caption{Uncurated random samples of our proposed robustness test sets Tiny ImageNetV2, -R and -A.}
        \label{fig:tin_samples}
\end{figure*}

\begin{figure*}[h]
    \centering
    \begin{subfigure}{0.99\textwidth}
         \centering
         \includegraphics[width=\textwidth]{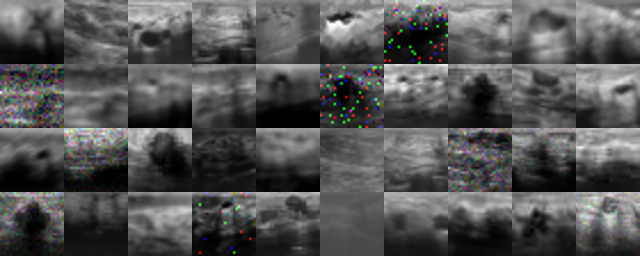}
         \caption{Uncurated random samples of BreastMNIST-C test set}
         \label{fig:breastmnist-c_samples}
    \end{subfigure}
    \hfill
    \\
    \begin{subfigure}{0.99\textwidth}
         \centering
         \includegraphics[width=\textwidth]{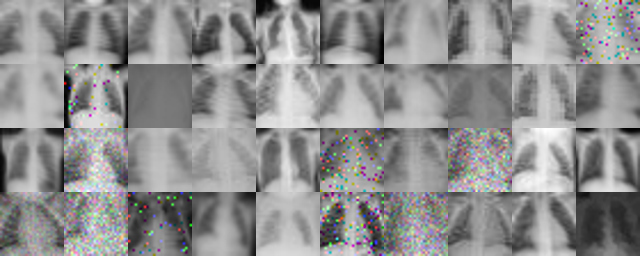}
         \caption{Uncurated random samples of PneumoniaMNIST-C test set}
         \label{fig:pneumoniamnist-c_samples}
    \end{subfigure}
    \hfill
    \\
    \begin{subfigure}{0.99\textwidth}
         \centering
         \includegraphics[width=\textwidth]{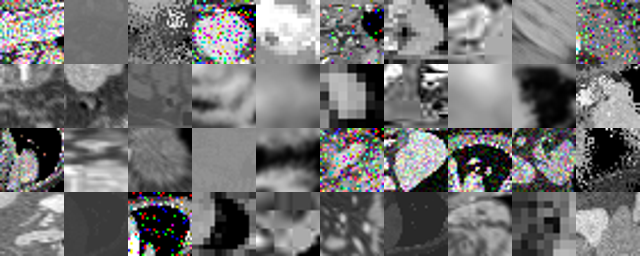}
         \caption{Uncurated random samples of OrganSMNIST test set}
         \label{fig:organsmnist-c_samples}
    \end{subfigure}
    \hfill
        \caption{Uncurated random samples of our proposed robustness test set collection MedMNIST-C.}
        \label{fig:medmnist_samples}
\end{figure*}

\begin{figure*}[h]
\centering
  \includegraphics[width=.88\linewidth]{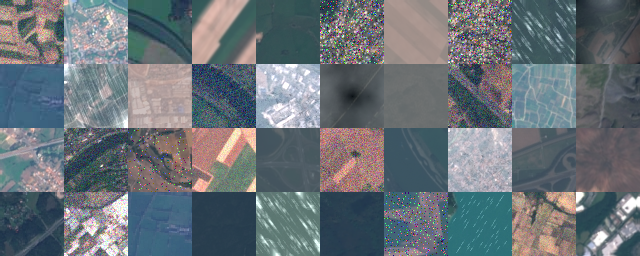}  
  \caption{Uncurated random samples of our proposed robustness aerial test set EuroSAT-C.}
  \label{fig:eurosat-c_samples}
\end{figure*}

In this section, we provide further information about the proposed Tiny ImageNetV2, -R and -A test sets.
In Table \ref{tab:datasets} we provide a summary of relevant statistics for all datasets.
This encompasses the total number of images, the amount of covered classes as well as channel-wise mean and standard deviations.
Furthermore, Figure \ref{fig:datasets} shows the distribution of the number of images available for each class.
In Figure \ref{fig:tin-r_class} it gets evident, that only 62 of 200 total classes are covered by Tiny-ImageNet-R.
However, apart from the missing classes in Tiny ImageNet-R, no significant anomaly becomes apparent in any of the datasets.
The class distribution of Tiny ImageNetV2 is not depicted as this test set is perfectly balanced covering all classes. Sample images of the proposed datasets are depicted in Figures \ref{fig:tin_samples} - \ref{fig:eurosat-c_samples}.

\begin{table*}[t]
\caption{Dataset details on Tiny ImageNetV2, -R and -A}
\label{tab:datasets}
\scriptsize
\centering
\begin{tabular}{lccc}
\hline
\multicolumn{1}{c}{\textbf{}} & \textbf{Tiny ImageNetV2} & \textbf{Tiny ImageNet-R}           & \textbf{Tiny ImageNet-A}           \\ \hline
Total Images                  & 2,000                    & 10,456                             & 3,374                              \\
Number of Classes             & 200 / 200                & 62 / 200                           & 200 / 200                          \\
Balanced?                     & \cmark                   & \xmark                             & \xmark                             \\
Min. Images per Class         & 10                       & 61                                 & 3                                  \\
Max. Images per Class         & 10                       & 430                                & 36                                 \\
Resolution                    & 64 $\times$ 64           & \multicolumn{1}{c}{64 $\times$ 64} & \multicolumn{1}{c}{64 $\times$ 64} \\
Mean (RGB)                    & (0.4705, 0.4415, 0.3913) & (0.6193, 0.5873, 0.5446)           & (0.4783, 0.4416, 0.3918)           \\
Std (RGB)                     & (0.2803, 0.2739, 0.2814) & (0.3162, 0.3176, 0.3356)           & (0.2781, 0.2711, 0.2796)           \\ \hline
\end{tabular}
\end{table*}

\clearpage

\end{document}